\documentclass{ieeeaccess}
\usepackage{cite}
\usepackage{amsmath,amssymb,amsfonts}
\usepackage{algorithmic}
\usepackage[pdftex]{graphicx}
\usepackage{bm}
\usepackage{algorithm}
\usepackage{algorithmic}
\usepackage{textcomp}
\def\BibTeX{{\rm B\kern-.05em{\sc i\kern-.025em b}\kern-.08em
    T\kern-.1667em\lower.7ex\hbox{E}\kern-.125emX}}
\begin{document}
\doi{10.1109/ACCESS.2022.3218216}

\title{Deep reinforcement learning under signal temporal logic constraints using Lagrangian relaxation}
\author{\uppercase{Junya Ikemoto}\authorrefmark{1} and
\uppercase{Toshimitsu Ushio\authorrefmark{2}} \IEEEmembership{Member, IEEE}.
}
\address{Graduate School of Engineering Science, Osaka University, Toyonaka 560-8531, Japan.} 
\address[1]{ikemoto@hopf.sys.es.osaka-u.ac.jp}
\address[2]{ushio@sys.es.osaka-u.ac.jp}
\tfootnote{This work was partially supported by JST CREST Grant Number JPMJCR2012, Japan and JSPS KAKENHI Grant Number JP21J10780, Japan.}

\markboth
{J. Ikemoto, T. Ushio: DRL Under STL Constraints Using Lagrangian Relaxation}
{J. Ikemoto, T. Ushio: DRL Under STL Constraints Using Lagrangian Relaxation}

\corresp{Corresponding author: Junya Ikemoto (e-mail: ikemoto@hopf.sys.es.osaka-u.ac.jp).}

\begin{abstract}
Deep reinforcement learning (DRL) has attracted much attention as an approach to solve optimal control problems without mathematical models of systems. On the other hand, in general, constraints may be imposed on optimal control problems. In this study, we consider the optimal control problems with constraints to complete temporal control tasks. We describe the constraints using signal temporal logic (STL), which is useful for time sensitive control tasks since it can specify continuous signals within bounded time intervals. To deal with the STL constraints, we introduce an extended constrained Markov decision process (CMDP), which is called a $\tau$-CMDP. We formulate the STL-constrained optimal control problem as the $\tau$-CMDP and propose a two-phase constrained DRL algorithm using the Lagrangian relaxation method. Through simulations, we also demonstrate the learning performance of the proposed algorithm.
\end{abstract}

\begin{keywords}
Constrained Reinforcement Learning, Deep Reinforcement Learning, Lagrangian Relaxation, Signal Temporal Logic.
\end{keywords}

\titlepgskip=-15pt

\maketitle

\section{Introduction}
\textit{Reinforcement learning} (RL) is a machine learning method for sequential decision making problems \cite{Sutton_RL}. In RL, a learner, which is called an \textit{agent}, interacts with an \textit{environment} and learns a desired policy automatically. Recently, RL with \textit{deep neural networks} (DNNs) \cite{DRL_book}, which is called \textit{Deep RL} (DRL), has attracted much attention for solving complicated decision making problems such as playing video games \cite{DQN}. DRL has been studied in various fields and many practical applications of DRL have been proposed \cite{DRL_manipulation,DRL_network,DRL_multi_agent}. On the other hand, when we apply RL or DRL to a problem in the real world, we must specify a state space of an environment for the problem beforehand. The states of the environment need to include sufficient information in order to determine a desired action at each time. Additionally, we must design a reward function for the task. If we do not design it to evaluate behaviors precisely, the learned policy may not be appropriate for the task.

Recently, controller design methods for temporal control tasks such as periodic, sequential, or reactive tasks have been studied in the control system community \cite{tl_control_system}. In these studies, \textit{linear temporal logic} (LTL) has often been used. LTL is one of temporal logics that have developed as formal methods in the computer science community \cite{TL}. LTL can express a temporal control task in a logical form. 

LTL has also been applied to RL for temporal control tasks \cite{Hasanbeig}. By using RL, we can obtain a policy to complete a temporal control task described by an LTL formula without a mathematical model of a system. The given LTL formula is transformed into an $\omega$-automaton that is a finite-state machine and accepts all traces satisfying the LTL formula. The transformed automaton can express states that include sufficient information to complete the temporal control task. We regard a system's state and an automaton's state as an environment's state for RL.The reward function for the temporal control task is designed based on the acceptance condition of the transformed automaton. Additionally, DRL algorithms for satisfying LTL formulae have been proposed in order to solve problems with continuous state-action spaces \cite{Yuan,Cai}. 

In real world problems, it is often necessary to describe temporal control tasks with time bounds. Unfortunately, LTL cannot express the time bounds. Then, \textit{metric interval temporal logic} (MITL) and \textit{signal temporal logic} (STL) are useful \cite{STL}. MITL is an extension of LTL and has time-constrained temporal operators. Furthermore, STL is an extension of MITL. Although LTL and MITL have predicates over Boolean signals, STL has inequality formed predicates over real-valued signals, which is useful to specify dynamical system's trajectories within bounded time intervals. Additionally, STL has a quantitative semantics called \textit{robustness} that evaluates how well a system's trajectory satisfies the given STL formula \cite{STL_Robustness}. In the control system community, controller design methods to complete tasks described by STL formulae have been proposed \cite{MPC_STL,CBF_STL}, where the control problems are formulated as constrained optimization problems using models of systems. Model-free RL-based controller design methods have also been proposed \cite{Aksaray,Venkataraman,Balakrishnan,Ikemoto_stl}. In \cite{Aksaray}, Aksaray \textit{et al}.\ proposed a \textit{Q-learning} algorithm for satisfying a given STL formula. The satisfaction of the given STL formula is based on a finite trajectory of the system. Thus, as an environment's state for a temporal control task, we use the extended state consisting of the current system's state and the previous system's states instead of using an automaton such as \cite{Hasanbeig}. Additionally, we design a reward function using the robustness for the given formula. In \cite{Venkataraman}, Venkataraman \textit{et al}.\ proposed a tractable learning method using a flag state instead of the previous system's state sequence to reduce the dimensionality of the environment's state space. However, these methods cannot be directly applied to problems with a continuous state-action space because they are based on a classical tabular Q-learning algorithm. For problems with continuous spaces, in \cite{Balakrishnan}, Balakrishnan \textit{et al}.\ introduced a \textit{partial signal} and applied a DRL algorithm to design a controller that partially satisfies a given STL specification and, in \cite{Ikemoto_stl}, we proposed a DRL-based design of a network controller to complete an STL control  task with network delays. 

On the other hand, for some control problems, we aim to design a policy that optimizes a given control performance index under a constraint described by an STL formula. For example, in practical applications, we should operate a system in order to satisfy a given STL formula with minimum fuel costs. In this study, we tackle to obtain the optimal policy for a given control performance index among the policies satisfying a given STL formula without a mathematical model of a system. 

\subsection{Contribution:} 
The main contribution is to propose a DRL algorithm to obtain an optimal policy for a given control performance index such as fuel costs under a constraint described by an STL formula. Our proposed algorithm has the following three advantages. 
\begin{enumerate}
\item We directly solve control problems with continuous state-action spaces. We apply DRL algorithms for problems with continuous spaces such as \textit{deep deterministic policy gradient} (DDPG) \cite{DDPG} and \textit{soft actor critic} (SAC) \cite{SAC}. 
\item We obtain a policy that not only satisfies a given STL formula but also is optimal with respect to a given control performance index. We consider the optimal control problem constrained by a given STL formula and formulate the problem as a \textit{constrained Markov decision process} (CMDP) \cite{CMDP}. In the CMDP problem, we introduce two reward functions: one is the reward function for the given control performance index and the other is the reward function for the given STL constraint. To solve the CMDP problem, we apply a constrained DRL (CDRL) algorithm with the \textit{Lagrangian relaxation} \cite{Model_free_CRL}. In this algorithm, we relax the CMDP problem into an unconstrained problem using a Lagrange multiplier to utilize standard DRL algorithms for problems with continuous spaces.  
\item We introduce a two-phase learning algorithm in order to make it easy to learn a policy satisfying the given STL formula. In a CMDP problem, it is important to satisfy the given constraint. The agent needs many experiences satisfying the given STL formula in order to learn how to satisfy the formula. However, it is difficult to collect the experiences considering both the control performance index and the STL constraint in the early learning stage since the agent may prioritize to optimize its policy with respect to the control performance index. Thus, in the first phase, the agent learns its policy without the control performance index in order to obtain experiences satisfying the STL constraint easily, which is called \textit{pre-training}. After obtaining many experiences satisfying the STL formula, in the second phase, the agent learns its optimal policy for the control performance index under the STL constraint, which is called \textit{fine-tuning}. 
\end{enumerate}
Through simulations, we demonstrate the learning performance of the proposed algorithm.

\subsection{Related works:}
\subsubsection{Classical RL for satisfying STL formulae}

Aksaray \textit{et al.} proposed a method to design policies satisfying STL formulae based on the Q-learning algorithm \cite{Aksaray}. However, in the method, the dimensionality of an environment's state tends to be large. Thus, Venkataraman \textit{et al.} proposed a tractable learning method to reduce the  dimensionality \cite{Venkataraman}. Furthermore, Kalagarla \textit{et al.}\ proposed an STL-constrained RL algorithm using a CMDP formulation and an online learning method \cite{Kalagarla}. However, since these are tabular-based approaches, we cannot directly apply them to problems with continuous spaces.  

\subsubsection{DRL for satisfying STL formulae}

DRL algorithms for satisfying STL formulae have been proposed \cite{Balakrishnan,Ikemoto_stl}. However, these studies focused on satisfying a given STL formula as the main objective. On the other hand, in this study, we regard the given STL formula as a constraint of a control problem and tackle the STL-constrained optimal control problem using a CDRL algorithm with the Lagrangian relaxation.   

\subsubsection{Learning with demonstrations for satisfying STL formulae}

Learning methods with demonstrations have been proposed \cite{Demo_stl_1,Demo_stl_2}. They designed a reward function using demonstrations, which was an imitation learning method. On the other hand, in this study, we do not use demonstrations to design a reward function for satisfying STL formulae. Alternatively, we design a reward function for satisfying STL formulae using robustness and the \textit{log-sum-exp} approximation \cite{Aksaray}.  

\subsection{Structure:} 
The remainder of this paper is organized as follows. In Section II, we review STL and the Q-learning algorithm to learn a policy satisfying STL formulae briefly. In Section III, we formulate an optimal control problem under a constraint described by an STL formula as a CMDP problem. In Section IV, we propose a CDRL algorithm with the Lagrangian relaxation to solve the CMDP problem. We relax the CMDP problem to an unconstrained problem using a Lagrange multiplier to utilize the DRL algorithm for unconstrained problems with continuous spaces. In Section V, by numerical simulations, we demonstrate the usefulness of the proposed method. In Section VI, we conclude the paper and discuss future works. 

\subsection{Notation:}
$\mathbb{N}_{\ge0}$ is the set of the nonnegative integers. $\mathbb{R}$ is the set of the real numbers. $\mathbb{R}_{\ge0}$ is the set of nonnegative real
numbers. $\mathbb{R}^{n}$ is the $n$-dimensional Euclidean space. For a set $A \subset \mathbb{R}$, $\max A$ and $\min A$ are the maximum and minimum value in $A$ if they exist, respectively. 
 
 \section{Preliminaries}
\subsection{Signal temporal logic}
We consider the following discrete-time stochastic dynamical system.
\begin{eqnarray}
x_{k+1} = f(x_{k},a_{k})+\Delta_{w} w_{k},\label{dynamics}
\end{eqnarray}
where $x_{k}\in\mathcal{X}$, $a_{k}\in\mathcal{A}$, and $w_{k}\in\mathcal{W}$ are the system's state, the agent's control action, and the system noise at $k\in\{0,1,...\}$. $\mathcal{X}=\mathbb{R}^{n_{x}}$, $\mathcal{A}\subseteq\mathbb{R}^{n_{a}}$, and $\mathcal{W}=\mathbb{R}^{n_{x}}$ are the system's state space, the control action space, and the system noise space, respectively. The system noise $w_{k}$ is an independent and identically distributed random variable with a probability density $p_{w}:\mathcal{W}\to\mathbb{R}_{\ge0}$. $\Delta_{w}$ is a regular matrix that is a weighting factor of the system noise. $f:\mathcal{X}\times\mathcal{A}\to\mathcal{X}$ is a function that describes the system dynamics. Then, we have the transition probability density $p_{f}(x'|x,a):=|\Delta_{w}^{-1}|p_{w}(\Delta_{w}^{-1}(x'-f(x,a)))$. The initial state $x_{0}\in\mathcal{X}$ is sampled from a probability density $p_{0}:\mathcal{X}\to\mathbb{R}_{\ge0}$. For a finite system trajectory whose length is $K+1$, $x_{k_1:k_2}$ denotes the partial trajectory for the time interval $[k_1,k_2]$, where $0\le k_{1} \le k_{2}\le K$. 

STL is a specification formalism that allows us to express real-time properties of real-valued trajectories of systems \cite{STL}. We consider the following \textit{syntax} of STL.  
\begin{eqnarray}
\Phi&::=&G_{[0,K_{e}]}\phi\ |\ F_{[0,K_{e}]}\phi,\nonumber\\
\phi&::=&G_{[k_{s},k_{e}]}\varphi\ |\ F_{[k_{s},k_{e}]}\varphi\ |\ \phi\land\phi\ |\ \phi\lor\phi,\nonumber\\
\varphi&::=&\psi\ |\ \lnot\varphi\ |\ \varphi\land\varphi\  |\ \varphi\lor\varphi,\nonumber
\end{eqnarray} 
where $K_{e}$, $k_{s}$, and $k_{e}\in\mathbb{N}_{\ge0}$ are nonnegative constants for the time bounds, $\Phi,\phi,\varphi,$ and $\psi$ are the STL formulae, $\psi$ is a predicate in the form of $h(x) \le d$, $h :\mathcal{X}\to\mathbb{R}$ is a function of the system's state, and $d\in\mathbb{R}$ is a constant. The Boolean operators $\lnot$, $\land$, and $\lor$ are negation, conjunction, and disjunction, respectively. The temporal operators $G_{\mathcal{T}}$ and $F_{\mathcal{T}}$ refer to \textit{Globally} (always) and \textit{Finally} (eventually), respectively, where $\mathcal{T}$ denotes the time bound of the temporal operator. $\phi_{i}=G_{[k_{s}^{i},k_{e}^{i}]}\varphi_{i}$, or $\phi_{i}=F_{[k_{s}^{i},k_{e}^{i}]}\varphi_{i}$, $i\in\{1,2,...,M\}$ are called \textit{STL sub-formulae}. $\phi$ comprises multiple STL sub-formulae $\{\phi_{i}\}_{i=1}^{M}$.

The \textit{Boolean semantics} of STL is recursively defined as follows:
\begin{eqnarray}
&& x_{k:K}\models\psi\Leftrightarrow h(x_{k})\le d,\nonumber\\
&& x_{k:K}\models\lnot\psi\Leftrightarrow \lnot(x_{k:K}\models\psi),\nonumber\\
&& x_{k:K}\models\phi_{1}\land\phi_{2}\Leftrightarrow x_{k:K}\models\phi_{1}\land x_{k:K}\models\phi_{2},\nonumber\\
&& x_{k:K}\models\phi_{1}\lor\phi_{2}\Leftrightarrow x_{k:K}\models\phi_{1}\lor x_{k:K}\models\phi_{2},\nonumber\\
&&x_{k:K}\models G_{[k_s,k_e]}\phi\Leftrightarrow\nonumber\\
&&\hspace{50pt}x_{k':K}\models\phi,\ \forall k'\in [k+k_s,k+k_e],\nonumber\\
&&x_{k:K}\models F_{[k_s,k_e]}\phi\Leftrightarrow\nonumber\\
&&\hspace{50pt}\exists k'\in [k+k_s,k+k_e],\ \text{s.t. } x_{k':K}\models\phi.\nonumber
\end{eqnarray}

The \textit{quantitative semantics} of STL, which is called robustness, is recursively defined as follows:
\begin{eqnarray}
\rho(x_{k:K},\psi)&=&d-h(x_{k}),\nonumber\\
\rho(x_{k:K},\lnot\psi)&=&-\rho(x_{k:K},\psi)\nonumber\\
\rho(x_{k:K},\phi_{1}\land\phi_{2})&=&\min\{\rho(x_{k:K},\phi_{1}),\rho(x_{k:K},\phi_{2})\},\nonumber\\
\rho(x_{k:K},\phi_{1}\lor\phi_{2})&=&\max\{\rho(x_{k:K},\phi_{1}),\rho(x_{k:K},\phi_{2})\},\nonumber\\
\rho(x_{k:K},G_{[k_s,k_e]}\phi)&=&\min_{k'\in [k+k_s,k+k_e]}\rho(x_{k':K},\phi),\nonumber\\
\rho(x_{k:K},F_{[k_s,k_e]}\phi)&=&\max_{k'\in [k+k_s,k+k_e]}\rho(x_{k':K},\phi),\nonumber
\end{eqnarray}
which quantifies how well the trajectory satisfies the given STL formulae \cite{STL_Robustness}.

The horizon length of an STL formula is recursively defined as follows:
\begin{eqnarray}
\text{hrz}(\psi)&=&0,\nonumber\\
\text{hrz}(\phi)&=& k_{e},\ \text{for }\phi=G_{[k_{s},k_{e}]}\varphi\ \text{or }F_{[k_s,k_e]}\varphi,\nonumber\\
\text{hrz}(\lnot\phi)&=&\text{hrz}(\phi),\nonumber\\
\text{hrz}(\phi_1\land\phi_2)&=&\max\{\text{hrz}(\phi_1),\text{hrz}(\phi_2)\},\nonumber\\
\text{hrz}(\phi_1\lor\phi_2)&=&\max\{\text{hrz}(\phi_1),\text{hrz}(\phi_2)\},\nonumber\\
\text{hrz}(G_{[k_s,k_e]}\phi)&=&k_e+\text{hrz}(\phi),\nonumber\\
\text{hrz}(F_{[k_s,k_e]}\phi)&=&k_e+\text{hrz}(\phi).\nonumber
\end{eqnarray}
$\text{hrz}(\phi)$ is the required length of the state sequence to verify the satisfaction of the STL formula $\phi$.

\subsection{Q-learning for Satisfying STL Formulae}
\begin{figure}[]
\begin{center}
  \includegraphics[width=8.0cm]{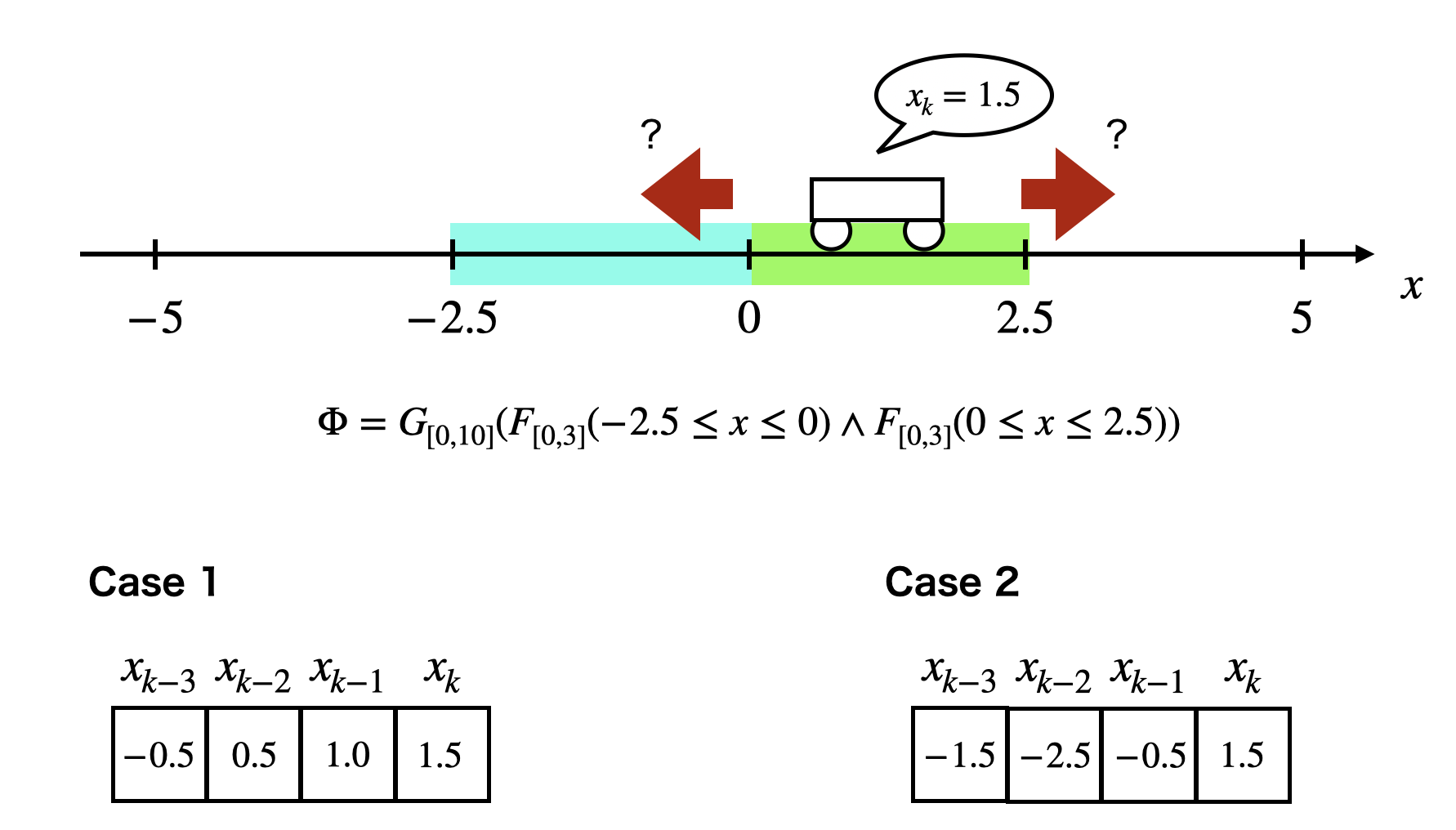}
  \caption{Illustration of a simple example of temporal control tasks described by STL formulae. }
  \label{simple_example_stl}
\end{center}
\end{figure}

In this section, we review the Q-learning algorithm to learn a policy satisfying a given STL formula \cite{Aksaray}. Although we often regard the current state of the dynamical system (\ref{dynamics}) as the environment's state for RL, the current system's state is not enough to determine an action for satisfying a given STL formula. Thus, Aksaray \textit{et al.} defined the following extended state using previous system's states.
\begin{eqnarray}
z_{k}=[x_{k-\tau+1}^{\top}\ x_{k-\tau+2}^{\top}\ ...\ x_{k}^{\top}]^{\top}\in\mathcal{Z},\nonumber
\end{eqnarray}  
where $\tau=\text{hrz}(\phi)+1$ for the given STL formula $\Phi=G_{[0,K_e]}\phi$ (or $\Phi=F_{[0,K_e]}\phi$) and $\mathcal{Z}$ is an extended state space.We show a simple example in Fig.\ \ref{simple_example_stl}. We operate a one-dimensional dynamical system to satisfy the STL formula 
\begin{eqnarray}
\Phi=G_{[0,10]}(F_{[0,3]}(-2.5\le x\le 0)\land F_{[0,3]}(0\le x\le 2.5)).\nonumber 
\end{eqnarray}
At any time in the time interval $[0,10]$, the system should enter both the blue  region and the green region before 3 time steps are elapsed, where there is no constraint for the order of the visits. Let the current system's state be $x_{k}=1.5$. Note that the desired  action for the STL formula is different depending on the past state sequence. For example, in the case where $x_{k-3:k}=-0.5,0.5,1.0,1.5$, we should operate the system to the blue region right away. On the other hand, in the case where $x_{k-3:k}=-1.5,-2.5,-0.5,1.5$, we do not need to move it. Thus, we regard not only the current system's state but also previous system's states as an environment's state for RL.          
Additionally, Aksaray \textit{et al.} designed the reward function $R_{STL}:\mathcal{Z}\to\mathbb{R}$ using robustness and the log-sum-exp approximation. The robustness of a trajectory $x_{0:K}$ with respect to the given STL formula $\Phi$ is as follows:
\begin{eqnarray}
&&\rho(x_{0:K},\Phi)\nonumber\\
&=&\begin{cases}
	\min\left\{\rho(x_{0:\tau-1},\phi),\ ...\ \rho(x_{K-\tau+1:K},\phi)\right\} \\ \hspace{36mm} \text{for }\Phi=G_{[0,K_e]}\phi,\\
	\max\left\{\rho(x_{0:\tau-1},\phi),\ ...\ \rho(x_{K-\tau+1:K},\phi)\right\} \\ \hspace{36mm} \text{for }\Phi=F_{[0,K_e]}\phi,\\
\end{cases}\nonumber\\
&=&\begin{cases}
	\min\left\{\rho(z_{\tau-1},\phi),\ ...\ \rho(z_{K},\phi)\right\} \\ \hspace{36mm} \text{for }\Phi=G_{[0,K_e]}\phi,\\
	\max\left\{\rho(z_{\tau-1},\phi),\ ...\ \rho(z_{K},\phi)\right\} \\ \hspace{36mm} \text{for }\Phi=F_{[0,K_e]}\phi.\\
\end{cases}
\end{eqnarray}
We consider the following problem.
\begin{eqnarray}
\max_{\pi}Pr\left[x_{0:K}^{\pi}\models \Phi\right]=\max_{\pi}E\left[\bm{1}(\rho(x_{0:K}^{\pi}, \Phi))\right],
\end{eqnarray}
where $x_{0:K}^{\pi}$ is the system's trajectory controlled by the policy $\pi$ and the function $\bm{1}:\mathbb{R}\to\{0,1\}$ is an indicator defined by
\begin{eqnarray}
\bm{1}(y)=\begin{cases}
	1 & \text{if }y\ge 0,\\
	0 & \text{if }y<0.
\end{cases}\label{indicator}
\end{eqnarray}
Since $\bm{1}(\min\{y_1,\ ...,\ y_n\})=\min\{\bm{1}(y_1),\ ...,\ \bm{1}(y_n)\}$ and $\bm{1}(\max\{y_1,\ ...,\ y_n\})=\max\{\bm{1}(y_1),\ ...,\ \bm{1}(y_n)\}$, 
\begin{eqnarray}
&&\max_{\pi}E\left[\bm{1}(\rho(x_{0:K}^{\pi}, \Phi))\right]\nonumber\\
&=&\begin{cases}
	\max_{\pi}E\left[\bm{1}(\min_{\tau-1\le k \le K}\rho(z_{k},\phi))\right] \\ \hspace{36mm} \text{for }\Phi=G_{[0,K_e]}\phi,\\
	\max_{\pi}E\left[\bm{1}(\max_{\tau-1\le k \le K}\rho(z_{k},\phi))\right] \\ \hspace{36mm} \text{for }\Phi=F_{[0,K_e]}\phi.\label{objective_Aksaray}
\end{cases}\nonumber\\
&=&\begin{cases}
	\max_{\pi}E\left[ \min_{\tau-1\le k \le K}\bm{1}(\rho(z_{k},\phi))\right] \\ \hspace{36mm} \text{for }\Phi=G_{[0,K_e]}\phi,\\
	\max_{\pi}E\left[ \max_{\tau-1\le k \le K}\bm{1}(\rho(z_{k},\phi))\right] \\ \hspace{36mm} \text{for }\Phi=F_{[0,K_e]}\phi.\label{objective_Aksaray}
\end{cases}
\end{eqnarray}
Then, we use the following log-sum-exp approximation. 
\begin{eqnarray}
\min\{y_1,\ ...,\ y_n\}&\simeq&-\frac{1}{\beta}\log\sum_{i=1}^{n}\exp(-\beta y_i),\label{min_log_sum_exp}\\
\max\{y_1,\ ...,\ y_n\}&\simeq&\frac{1}{\beta}\log\sum_{i=1}^{n}\exp(\beta y_i),\label{max_log_sum_exp}
\end{eqnarray}
where $\beta>0$ is an approximation parameter. We can approximate $\min\{\cdots\}$ or $\max\{\cdots\}$ with arbitrary accuracy by selecting a large $\beta$. Then, (\ref{objective_Aksaray}) can be approximated as follows:
\begin{eqnarray}
&&\max_{\pi}E[\bm{1}(\rho(x_{0:K}^{\pi}, \Phi))]\nonumber\\
&\simeq&\begin{cases}
	\max_{\pi}E\left[ -\frac{1}{\beta}\log\sum_{k=\tau-1}^{K}\exp(-\beta\bm{1}(\rho(z_{k},\phi)))\right] \\ \hspace{45mm} \text{for }\Phi=G_{[0,K_e]}\phi,\\
	\max_{\pi}E\left[ \frac{1}{\beta}\log\sum_{k=\tau-1}^{K}\exp(\beta\bm{1}(\rho(z_{k},\phi)))\right] \\ \hspace{45mm} \text{for }\Phi=F_{[0,K_e]}\phi.
\end{cases}\nonumber
\end{eqnarray}
Since the $\log$ function is a strictly monotonic function and $\beta>0$ is a constant, we have
\begin{eqnarray}
\begin{cases}
	\max_{\pi}E\left[ -\frac{1}{\beta}\log\sum_{k=\tau-1}^{K}\exp(-\beta\bm{1}(\rho(z_{k},\phi)))\right] \\ \hspace{45mm} \text{for }\Phi=G_{[0,K_e]}\phi,\\
	\max_{\pi}E\left[ \frac{1}{\beta}\log\sum_{k=\tau-1}^{K}\exp(\beta\bm{1}(\rho(z_{k},\phi)))\right] \\ \hspace{45mm} \text{for }\Phi=F_{[0,K_e]}\phi.
\end{cases}\nonumber\\
\Leftrightarrow\begin{cases}
	\max_{\pi}E\left[\sum_{k=\tau-1}^{K}-\exp(-\beta\bm{1}(\rho(z_{k},\phi)))\right] \\ \hspace{45mm} \text{for }\Phi=G_{[0,K_e]}\phi,\\
	\max_{\pi}E\left[\sum_{k=\tau-1}^{K}\exp(\beta\bm{1}(\rho(z_{k},\phi)))\right] \\ \hspace{45mm} \text{for }\Phi=F_{[0,K_e]}\phi.\nonumber
\end{cases}
\end{eqnarray}
Thus, we use the following reward function $R_{STL}:\mathcal{Z}\to\mathbb{R}$ to satisfy the given STL formula $\Phi$.
\begin{eqnarray}
&R_{STL}(z)&\nonumber\\
&=&\begin{cases}
	-\exp(-\beta\bm{1}(\rho(z,\phi))) & \text{if }\Phi=G_{[0,K_e]}\phi,\\
	\exp(\beta\bm{1}(\rho(z,\phi))) & \text{if }\Phi=F_{[0,K_e]}\phi.
\end{cases}\nonumber\\
\label{tau_reward}
\end{eqnarray}

\begin{figure}[h]
\begin{center}
  \includegraphics[width=7.5cm]{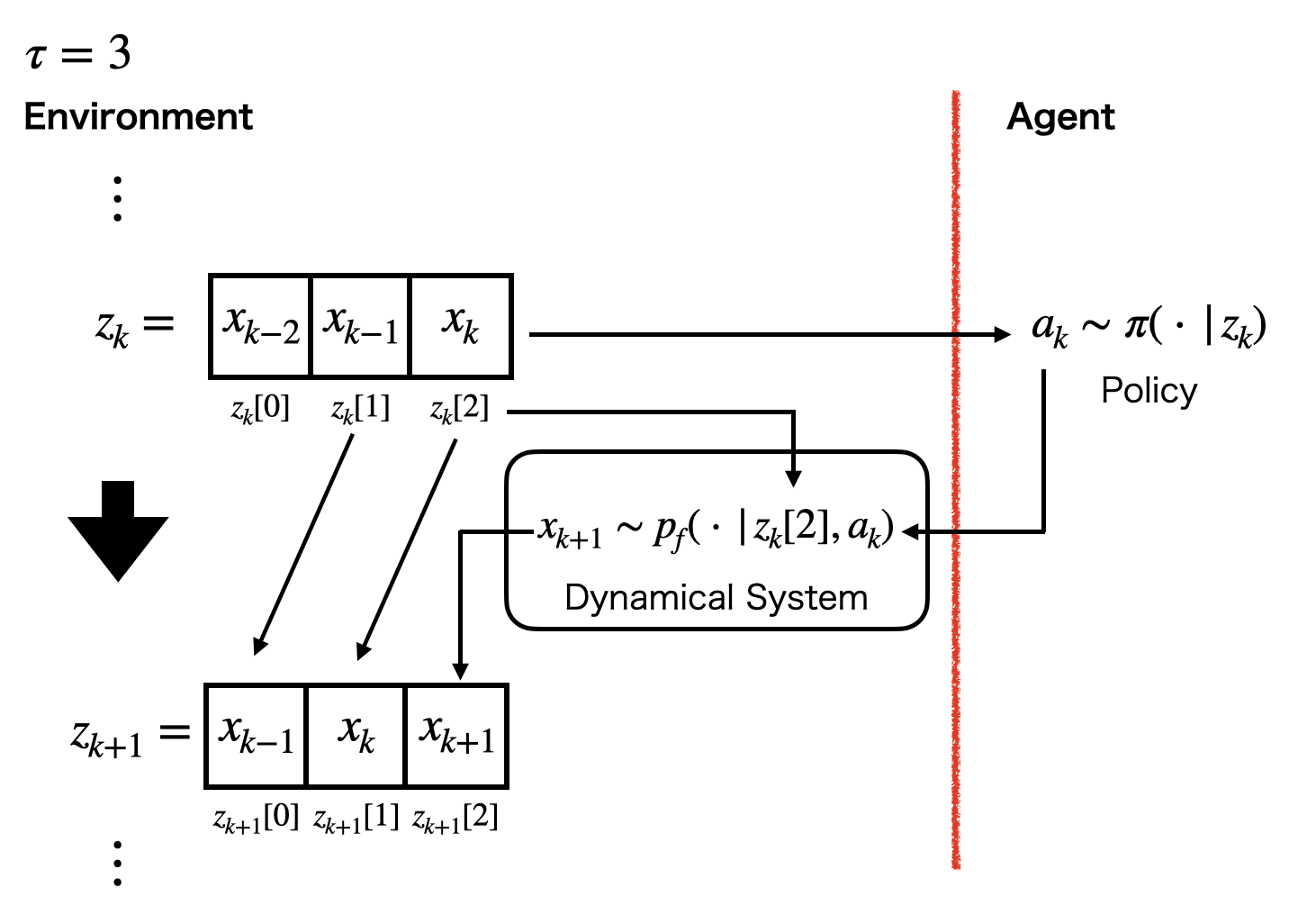}
  \caption{Illustration of an extended state transition. We consider the case $\tau=3$. The next extended state $z_{k+1}$ depends on the current extended state $z_k$ and the agent's action $a_k$. }
  \label{tau_MDP}
\end{center}
\end{figure}

To design a controller satisfying an STL formula using the Q-learning algorithm, Aksaray \textit{et al.} proposed a $\tau$-MDP as follows: 

\noindent \textbf{Definition 1 ($\tau$-MDP):}
We consider an STL formula $\Phi=G_{[0,K_{e}]}\phi$ (or $\Phi=F_{[0,K_{e}]}\phi$), where $\text{hrz}(\Phi)=K$ and $\phi$ comprises multiple STL sub-formulae $\phi_{i},\ i\in\{1,2,...,M\}$. Subsequently, we set $\tau=\text{hrz}(\phi)+1$, that is, $K=K_e+\tau-1$. A $\tau$-MDP is defined by a tuple $\mathcal{M}_{\tau}=\left<\mathcal{Z},\mathcal{A},p_{0}^{z},p^{z},R_{STL}\right>$, where
\begin{itemize}
\item $\mathcal{Z}\subseteq\mathcal{X}^{\tau}$ is an extended state space that is an environment's state space for RL. The extended state $z\in\mathcal{Z}$ is a vector of multiple system's states $z=[z[0]^{\top}\ z[1]^{\top}\ ...\ z[\tau-1]^{\top}]^{\top},\ z[i]\in\mathcal{X},\ \forall i\in\{0,1,...,\tau-1\}$.
\item $\mathcal{A}$ is an agent's control action space.
\item $p_{0}^{z}$ is a probability density for the initial extended state $z_{0}$ with $z_{0}[i]=x_{0},\ \forall i\in\{0,1,...,\tau-1\}$, where $x_{0}$ is generated from $p_{0}$.
\item $p^{z}$ is a transition probability density for the extended state. When the system's state is updated by $x'\sim p_{f}(\cdot |x,a)$, the extended state is updated by $z'\sim p^{z}(\cdot|z,a)$ as follows:
\begin{eqnarray}
z'[i]&=&z[i+1],\ \forall i\in\{0,1,..,\tau-2\},\nonumber\\
z'[\tau-1]&\sim&p_{f}(\cdot|z[\tau-1],a).\nonumber
\end{eqnarray}
Fig. \ref{tau_MDP} shows an example of the transition. We consider the sequence that consists of $\tau$ system's states $x_{k-\tau+1},x_{k-\tau+2},...,x_{k}$ as the extended state at time $k$. In the transition, the head system's state $x_{k-\tau+1}$ is removed from the sequence and other system's states $x_{k-\tau+2},...,x_{k}$ are shifted to the left. After that, the next system's state $x_{k+1}$ updated by $p_{f}(\cdot|x_{k},a_{k})$ is inputted to the tail of the sequence. The next extended state $z_{k+1}$ depends on the current extended state $z_{k}$ and the agent's action $a_k$.
\item $R_{STL}:\mathcal{Z}\to\mathbb{R}$ is the \textit{STL-reward function} defined by (\ref{tau_reward}).
\end{itemize}

\section{Problem formulation}
We consider the following optimal policy design problem constrained by a given STL formula $\Phi$, where the system model (\ref{dynamics}) is unknown.  
\begin{equation}
\begin{split}
&\text{maximize}_{\pi}\ E_{p_\pi}\left[\sum_{k=0}^{K}\gamma^{k}R(x_{k},a_{k})\right],\\
&\text{subject to}\ \ \ \ x_{0:K}\models\Phi,
\end{split}\label{main_problem}
\end{equation}
where $\gamma\in[0,1)$ is a discount factor and $R:\mathcal{X}\times\mathcal{A}\to\mathbb{R}$ is a reward function for a given control performance index. $E_{p_\pi}[\cdot]$ is the expectation value with respect to the distributions $p_{0}$, $p_f$, and $\pi$. We introduce the following $\tau$-\textit{CMDP} that is an extension of a $\tau$-MDP \cite{Aksaray} to deal with the problem (\ref{main_problem}).

\noindent \textbf{Definition 2 ($\tau$-CMDP):}
We consider an STL formula $\Phi=G_{[0,K_{e}]}\phi$ (or $\Phi=F_{[0,K_{e}]}\phi$) as a constraint, where $\text{hrz}(\Phi)=K$ and $\phi$ comprises multiple STL sub-formulae $\phi_{i},\ i\in\{1,2,...,M\}$. Subsequently, we set $\tau=\text{hrz}(\phi)+1$, that is, $K=K_e+\tau-1$. A $\tau$-CMDP is defined by a tuple $\mathcal{CM}_{\tau}=\left<\mathcal{Z},\mathcal{A},p_{0}^{z},p^{z},R_{STL},R_{z}\right>$, where
\begin{itemize}
\item $\mathcal{Z}\subseteq\mathcal{X}^{\tau}$ is an extended state space that is an environment's state space for RL. The extended state $z\in\mathcal{Z}$ is a vector of multiple system's states $z=[z[0]^{\top}\ z[1]^{\top}\ ...\ z[\tau-1]^{\top}]^{\top},\ z[i]\in\mathcal{X},\ \forall i\in\{0,1,...,\tau-1\}$.
\item $\mathcal{A}$ is an agent's control action space.
\item $p_{0}^{z}$ is a probability density for the initial extended state $z_{0}$ with $z_{0}[i]=x_{0},\ \forall i\in\{0,1,...,\tau-1\}$, where $x_{0}$ is generated from $p_{0}$.
\item $p^{z}$ is a transition probability density for the extended state. When the system's state is updated by $x'\sim p_{f}(\cdot |x,a)$, the extended state is updated by $z'\sim p^{z}(\cdot|z,a)$ as follows:
\begin{eqnarray}
z'[i]&=&z[i+1],\ \forall i\in\{0,1,..,\tau-2\},\nonumber\\
z'[\tau-1]&\sim&p_{f}(\cdot|z[\tau-1],a).\nonumber
\end{eqnarray}
\item $R_{STL}:\mathcal{Z}\to\mathbb{R}$ is the STL-reward function defined by (\ref{tau_reward}) for satisfying the given STL formula $\Phi$.
\item $R_{z}:\mathcal{Z}\times\mathcal{A}\to\mathbb{R}$ is a reward function as follows:
\begin{eqnarray}
R_{z}(z,a)=R(z[\tau-1],a),\nonumber
\end{eqnarray}
where $R:\mathcal{X}\times\mathcal{A}\to\mathbb{R}$ is a reward function for a given control performance index. 
\end{itemize} 

We design an optimal policy with respect to $R_{z}$ under satisfying the STL formula using a model-free CDRL algorithm \cite{Model_free_CRL}. Then, we define the following functions.
\begin{eqnarray}
J(\pi)&=&E_{p_\pi}\left[\sum_{k=0}^{K}\gamma^{k}R_{z}(z_{k},a_{k})\right],\nonumber\\
J_{STL}(\pi)&=&E_{p_\pi}\left[\sum_{k=0}^{K}\gamma^{k}R_{STL}(z_{k})\right],\nonumber
\end{eqnarray}
where $\gamma\in[0,1)$ is a discount factor close to $1$. $E_{p_\pi}[\cdot]$ is the expectation value with respect to the distributions $p_{0}$, $p_f$, and $\pi$. We reformulate the problem (\ref{main_problem}) as follows:
\begin{eqnarray}
&&\pi^{*}\in\arg\max_{\pi}\ J(\pi),\label{objective}\\
&&\text{subject to}\ \ J_{STL}(\pi)\ge l_{STL},\label{constraint}
\end{eqnarray}
where $l_{STL}\in\mathbb{R}$ is a lower threshold. In this study, $l_{STL}$ is a hyper-parameter for adjusting the satisfiability of the given STL formula. The larger $l_{STL}$ is, the more conservatively the agent learns a policy to satisfy the STL formula. We call the constrained problem with (\ref{objective}) and (\ref{constraint}) a $\tau$-\textit{CMDP problem}. In the next section, we propose a CDRL algorithm with the Lagrangian relaxation to solve the $\tau$-CMDP problem.  

\section{Deep reinforcement learning under a signal temporal logic constraint}
We propose a CDRL algorithm with the Lagrangian relaxation to obtain an optimal policy for the $\tau$-CMDP problem. Our proposed algorithm is based on the DDPG algorithm \cite{DDPG} or the SAC algorithm \cite{SAC}, which are DRL algorithms derived from the Q-learning algorithm for problems with continuous state-action spaces. In both algorithms, we parameterize an agent's policy $\pi$ using a DNN, which is called an \textit{actor DNN}. The agent updates the parameter vector of the actor DNN based on $J(\pi)$. However, in this problem, the agent cannot directly use $J(\pi)$ since the mathematical model of the system $p_f$ is unknown. Thus, we approximate $J(\pi)$ using another DNN, which is called a \textit{critic DNN}. Additionally, we use the following two techniques proposed in \cite{DQN}.
\begin{itemize}
\item \textit{Experience replay}, 
\item \textit{Target network}.
\end{itemize}
In the experience replay, the agent does not update the parameter vectors of DNNs immediately when obtaining an experience. Alternatively, the agent stores the obtained experience to the \textit{replay buffer} $\mathcal{D}$. The agent selects some experiences from the replay buffer $\mathcal{D}$ randomly and updates the parameter vector of DNNs using the selected experiences. The experience replay can reduce correlation among experience data. In the target network technique, we prepare separate DNNs for the critic DNN and the actor DNN, which are called a \textit{target critic DNN} and a \textit{target actor DNN}, respectively, and output target values for updates of the critic DNN. The parameter vectors of the target DNNs are updated by tracking the parameter vectors of the actor DNN and the critic DNN slowly. If we do not use the target network technique for updates of the critic DNN, we need to compute the target value using the current critic DNN, which is called \textit{bootstrapping}. If we update the critic DNN substantially, the target value computed by the updated critic DNN may change largely, which leads to oscillations of the learning performance. It is known that the target network technique can improve the learning stability.   

\noindent \textbf{Remark:} The standard DRL algorithm based on Q-learning is the DQN algorithm \cite{DQN}. However, the DQN algorithm cannot handle continuous action spaces due to its DNN architecture.

On the other hand, we cannot directly apply the DDPG algorithm and the SAC algorithm to the $\tau$-CMDP problem since these are algorithms for unconstrained problems. Thus, we consider the following \textit{Lagrangian relaxation} \cite{Constrained_Optimization}.
\begin{eqnarray}
\min_{\kappa\ge0}\max_{\pi}\mathcal{L}(\pi,\kappa),\label{Lagrangian_Problem}
\end{eqnarray}
where $\mathcal{L}(\pi,\kappa)$ is a \textit{Lagrangian function} given by 
\begin{eqnarray}
\mathcal{L}(\pi,\kappa)=J(\pi)+\kappa(J_{STL}(\pi)-l_{STL}),\label{Lagrangian}
\end{eqnarray}
and $\kappa\ge0$ is a \textit{Lagrange multiplier}. We can relax the constrained problem into the unconstrained problem.

\begin{figure}[h]
\begin{center}
  \includegraphics[width=7.5cm]{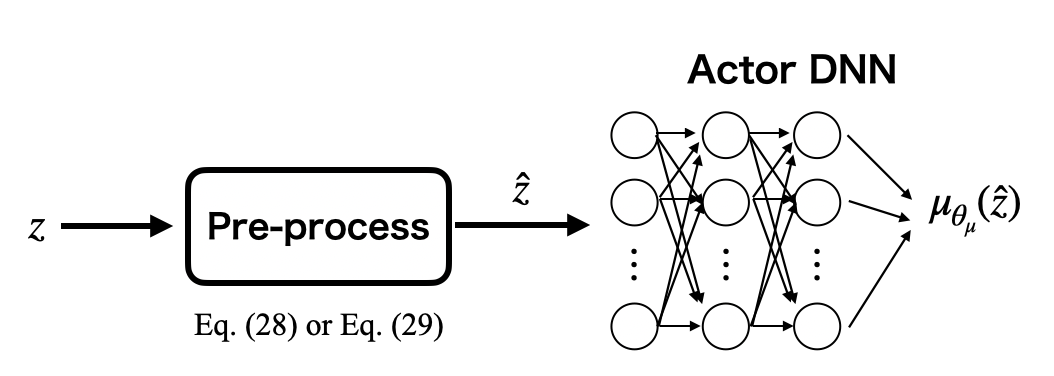}
  \caption{Illustration of an actor DNN for the DDPG-Lagrangian algorithm. Actually, we input a pre-processed state $\hat{z}$ stated in Section IV.D to the DNN instead of an extended state $z$.}
  \label{DDPG_actor}
\end{center}
\end{figure}

\begin{figure}[h]
\begin{center}
  \includegraphics[width=8.5cm]{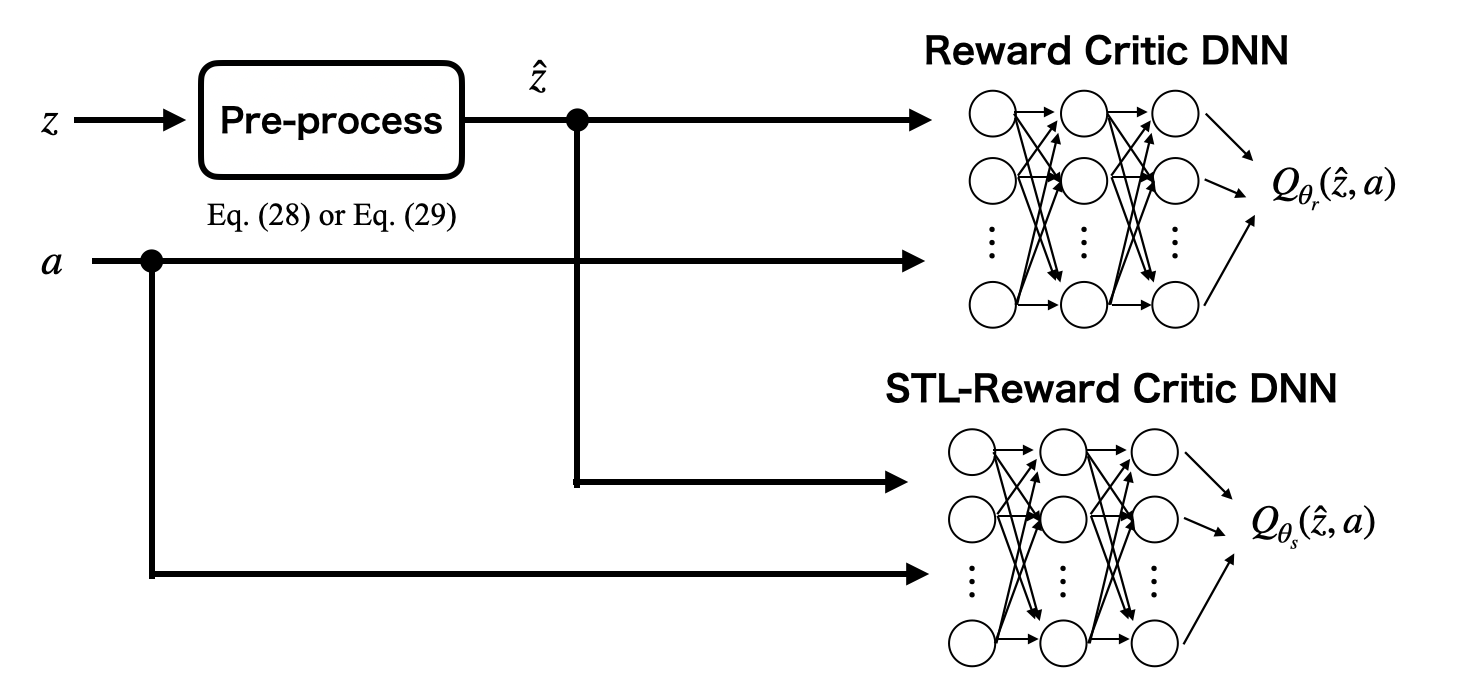}
  \caption{Illustration of the two-type critic DNNs (the reward critic DNN and the STL-reward critic DNN). In the DDPG-Lagrangian algorithm, the reward critic DNN and the STL-reward critic estimate the terms $J(\mu_{\theta_{\mu}})$ and $J_{STL}(\mu_{\theta_{\mu}})$ in (\ref{Lagrangian}), respectively. In the SAC-Lagrangian algorithm, the reward critic DNN and the STL-reward critic DNN estimate the terms $J_{ent}(\pi_{\theta_{\pi}})$ and $J_{STL}(\pi_{\theta_{\pi}})$ in (\ref{SAC_Lagrangian}), respectively. Actually, we input a pre-processed state $\hat{z}$ stated in Section IV.D to the DNN instead of an extended state $z$.}
  \label{Critic_DNN}
\end{center}
\end{figure}

\subsection{DDPG-Lagrangian}
We parameterize a deterministic policy using a DNN as shown in Fig.\ \ref{DDPG_actor}, which is an actor DNN. Its parameter vector is denoted by $\theta_{\mu}$. In the DDPG-Lagrangian algorithm, the parameter vector $\theta_{\mu}$ is updated by maximizing (\ref{Lagrangian}). However, $J(\mu_{\theta_{\mu}})$ and $J_{STL}(\mu_{\theta_{\mu}})$ are unknown. Thus, as shown in Fig.\ \ref{Critic_DNN}, $J(\mu_{\theta_{\mu}})$ and $J_{STL}(\mu_{\theta_{\mu}})$ are approximated by two separate critic DNNs, which are called a \textit{reward critic DNN} and an \textit{STL-reward critic DNN}, respectively. The parameter vectors of the reward critic DNN and the STL-reward critic DNN are denoted by $\theta_{r}$ and $\theta_{s}$, respectively. $\theta_{r}$ and $\theta_{s}$ are updated by decreasing the following critic loss functions.
\begin{eqnarray}
J_{rc}(\theta_{r})&=&E_{(z,a,z')\sim\mathcal{D}}\left[\left(Q_{\theta_{r}}(z,a)-t_r\right)^2\right],\label{reward_DDPG_critic_update}\\
J_{sc}(\theta_{s})&=&E_{(z,a,z')\sim\mathcal{D}}\left[\left(Q_{\theta_{s}}(z,a)-t_s\right)^2\right],\label{STL_DDPG_reward_critic_update}
\end{eqnarray}
where $Q_{\theta_{r}}(\cdot,\cdot)$ and $Q_{\theta_{s}}(\cdot,\cdot)$ are the outputs of the reward critic DNN and the STL-reward critic DNN, respectively. The target values $t_r$ and $t_s$ are given by 
\begin{eqnarray}
t_r&=&R_{z}(z,a)+\gamma Q_{\theta_{r}^{-}}(z',\mu_{\theta_{\mu}^{-}}(z')),\nonumber\\
t_s&=&R_{STL}(z)+\gamma Q_{\theta_{s}^{-}}(z',\mu_{\theta_{\mu}^{-}}(z')).\nonumber
\end{eqnarray}
$Q_{\theta_{r}^{-}}(\cdot,\cdot)$ and $Q_{\theta_{s}^{-}}(\cdot,\cdot)$ are the outputs of the target reward critic DNN and the target STL-reward critic DNN, respectively, and $\mu_{\theta_{\mu}^{-}}(\cdot)$ is the output of target actor DNN. $\theta_{r}^{-}$, $\theta_{s}^{-}$, and $\theta_{\mu}^{-}$ are parameter vectors of the target reward critic DNN, the target STL-reward critic DNN, and the target actor DNN, respectively. Their parameter vectors are slowly updated by the following \textit{soft update}.
\begin{eqnarray}
\theta_{r}^{-}&\leftarrow& \xi\theta_{r}+(1-\xi)\theta_{r}^{-},\nonumber\\
\theta_{s}^{-}&\leftarrow& \xi\theta_{s}+(1-\xi)\theta_{s}^{-},\label{soft_update}\\
\theta_{\mu}^{-}&\leftarrow& \xi\theta_{\mu}+(1-\xi)\theta_{\mu}^{-},\nonumber
\end{eqnarray}
where $\xi>0$ is a sufficiently small positive constant. The agent stores experiences to the replay buffer $\mathcal{D}$ and selects some experiences from $\mathcal{D}$ randomly for updates of $\theta_{r}$ and $\theta_{s}$. $E_{(z,a,z')\sim\mathcal{D}}[\cdot]$ is the expected value under the random sampling of the experiences from $\mathcal{D}$. In the standard DDPG algorithm \cite{DDPG}, the parameter vector of the actor DNN is updated by decreasing 
\begin{eqnarray}
J_{a}(\theta_{\mu})=E_{z\sim\mathcal{D}}[-Q_{\theta_{r}}(z,\mu_{\theta_{\mu}}(z))],\nonumber
\end{eqnarray}
where $E_{z\sim\mathcal{D}}[\cdot]$ is the expected value with respect to $z$ sampled from $\mathcal{D}$ randomly. However, in the DDPG-Lagrangian algorithm, we consider (\ref{Lagrangian}) as an objective instead of $J(\mu_{\theta_{\mu}})$. Thus, the parameter vector of the actor DNN $\theta_{\mu}$ is updated by decreasing the following actor loss function.
\begin{eqnarray}
&&J_{a}(\theta_{\mu})=\nonumber\\
&&\ \ \ E_{z\sim\mathcal{D}}[-(Q_{\theta_{r}}(z,\mu_{\theta_{\mu}}(z))+\kappa Q_{\theta_{s}}(z,\mu_{\theta_{\mu}}(z)))].\nonumber\\
\label{DDPG_actor_update}
\end{eqnarray}
The Lagrange multiplier $\kappa$ is updated by decreasing the following loss function.
\begin{eqnarray}
J_{L}(\kappa)=E_{z_0\sim p_{0}^{z}}\left[\kappa(Q_{\theta_{s}}(z_0,\mu_{\theta_{\mu}}(z_0))-l_{STL})\right],\label{DDPG_kappa_update}
\end{eqnarray}
where $E_{z_0\sim p_{0}^{z}}[\cdot]$ is the expected value with respect to $p_{0}^{z}$. 

\noindent \textbf{Remark:} $\kappa$ is a nonnegative parameter adjusting the relative importance of the STL-reward critic DNN against the reward critic DNN in updating the actor DNN. Intuitively, if the agent's policy does not satisfy (\ref{constraint}), then we increase the parameter $\kappa$, which increases the relative importance of the STL-critic DNN. On the other hand, if the agent's policy satisfies (\ref{constraint}), then we decrease the parameter $\kappa$, which decreases the relative importance of the STL-critic DNN. 

\begin{figure*}[h]
\begin{center}
  \includegraphics[width=14cm]{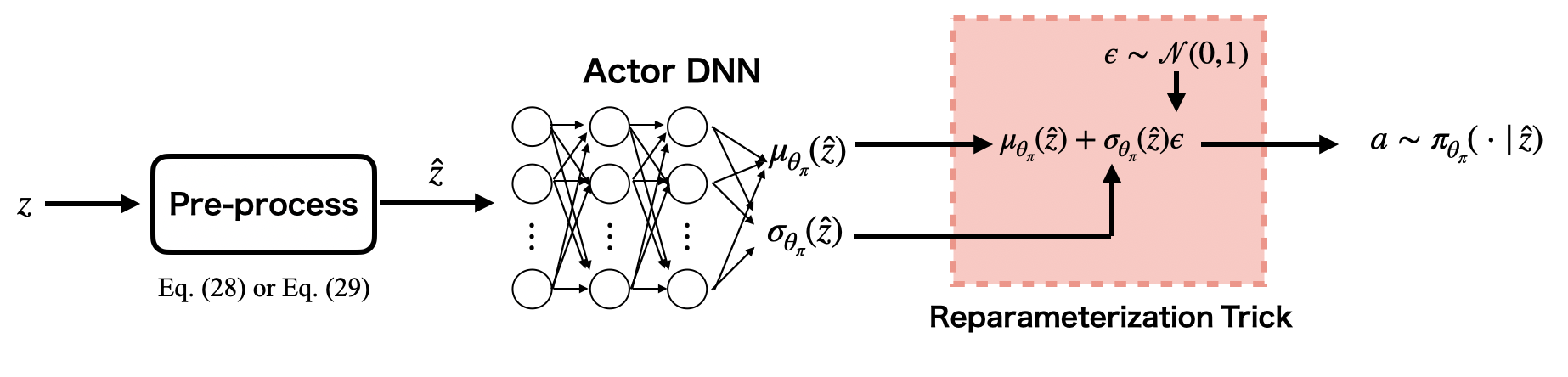}
  \caption{Illustration of an actor DNN with a reparameterization trick. The DNN outputs the mean $\mu_{\theta_{\pi}}(\hat{z})$ and the standard deviation $\sigma_{\theta_{\pi}}(\hat{z})$ parameters for an input $\hat{z}$. We use the reparameterization trick to sample an action, where $\epsilon$ is sampled from a standard normal distribution $\mathcal{N}(0,1)$. }
  \label{Actor_DNN}
\end{center}
\end{figure*}

\subsection{SAC-Lagrangian}
SAC is a maximum entropy DRL algorithm that obtains a policy to maximize both the expected sum of rewards and the expected entropy of the policy. It is known that a maximum entropy algorithm  improves explorations by acquiring diverse behaviors and has the robustness for the estimation error \cite{SAC}. In the SAC algorithm, we design a stochastic policy $\pi$. We use the following objective with an entropy term instead of $J(\pi)$. 
\begin{eqnarray}
J_{ent}(\pi)&=&E_{p_\pi}\left[\sum_{k=0}^{K}\gamma^{k}(R_{z}(z_{k},a_{k})+\alpha\mathcal{H}(\pi(\cdot|z_{k})))\right],\nonumber\\
&=&J(\pi)+E_{p_\pi}\left[\sum_{k=0}^{K}\gamma^{k}\alpha\mathcal{H}(\pi(\cdot|z_{k}))\right],
\label{ent_objective}
\end{eqnarray}
where $\mathcal{H}(\pi(\cdot|z_{k}))=E_{a\sim\pi}[-\log\pi(a|z_{k})]$ is an entropy of the stochastic policy $\pi$ and $\alpha\ge0$ is an entropy temperature. The entropy temperature determines the relative importance of the entropy term against the sum of rewards. 

We use the Lagrangian relaxation for the SAC algorithm such as  \cite{SAC_Lagrangian,CSAC}. Then, a Lagrangian function with the entropy term is given by 
\begin{eqnarray}
\mathcal{L}(\pi,\kappa)=J_{ent}(\pi)+\kappa(J_{STL}(\pi)-l_{STL}).\label{SAC_Lagrangian}
\end{eqnarray}
We model the stochastic policy $\pi_{\theta_{\pi}}$ using a Gaussian with the mean and the standard deviation outputted by a DNN with a \textit{reparameterization trick} \cite{VAE} as shown in Fig.\ \ref{Actor_DNN}, which is an actor DNN. The parameter vector is denoted by $\theta_{\pi}$. Additionally, we need to estimate $J_{ent}(\pi_{\theta_{\pi}})$ and $J_{STL}(\pi_{\theta_{\pi}})$ to update the parameter vector $\theta_{\pi}$ like the DDPG-Lagrangian algorithm. Thus, $J_{ent}(\pi_{\theta_{\pi}})$ and $J_{STL}(\pi_{\theta_{\pi}})$ are also approximated by two separate critic DNNs as shown in Fig.\ \ref{Critic_DNN}. Note that, in the SAC-Lagrangian algorithm, the reward critic DNN estimates not only $J(\pi_{\theta_{\pi}})$ but also the entropy term. The parameter vectors are also updated using the experience replay and the target network technique. $\theta_{r}$ and $\theta_{s}$ are updated by decreasing the following critic loss functions.
\begin{eqnarray}
&&J_{rc}(\theta_{r})=\nonumber\\
&&\ \ \ E_{(z,a,z')\sim\mathcal{D}}\left[\left(Q_{\theta_{r}}(z,a)-\left(r+\gamma V_{\theta_{r}^{-}}(z')\right)\right)^2\right],\nonumber\\
\label{reward_critic_update}\\
&&J_{sc}(\theta_{s})=\nonumber\\
&&\ \ \ E_{(z,a,z')\sim\mathcal{D}}\left[\left(Q_{\theta_{s}}(z,a)-\left(s+\gamma V_{\theta_{s}^{-}}(z')\right)\right)^2\right],\nonumber\\
\label{STL_reward_critic_update}
\end{eqnarray}
where $r=R_{z}(z,a)$, $s=R_{STL}(z)$, and $Q_{\theta_{r}}(\cdot,\cdot)$ and $Q_{\theta_{s}}(\cdot,\cdot)$ are the outputs of the reward critic DNN and the STL-reward critic DNN, respectively. The target values are computed by 
\begin{eqnarray}
V_{\theta_{r}^{-}}(z')&=&E_{a'\sim\pi_{\theta_{\pi}}}\left[Q_{\theta_{r}^{-}}(z',a')-\alpha\log\pi_{\theta_{\pi}}(a'|z')\right],\nonumber\\
V_{\theta_{s}^{-}}(z')&=&E_{a'\sim\pi_{\theta_{\pi}}}\left[Q_{\theta_{s}^{-}}(z',a')\right],\nonumber
\end{eqnarray}
where $Q_{\theta_{r}^{-}}(\cdot,\cdot)$ and $Q_{\theta_{s}^{-}}(\cdot,\cdot)$ are outputs of the target reward critic DNN and the target STL-reward critic DNN, respectively, and $E_{a'\sim\pi_{\theta_{\pi}}}[\cdot]$ is the expected value with respect to $\pi_{\theta_{\pi}}$. Their parameter vectors $\theta_{r}^{-}$, $\theta_{s}^{-}$ are slowly updated like (\ref{soft_update}). In the standard SAC algorithm, the parameter vector of the actor DNN $\theta_{\pi}$ is updated by decreasing 
\begin{eqnarray}
J_{a}(\theta_{\pi})=E_{z\sim\mathcal{D},a\sim\pi_{\theta_{\pi}}}[\alpha\log(\pi_{\theta_{\pi}}(a|z))-Q_{\theta_{r}}(z,a)],\nonumber
\end{eqnarray}
where $E_{z\sim\mathcal{D},a\sim\pi_{\theta_{\pi}}}[\cdot]$ is the expected value with respect to the experiences $z$ sampled from $\mathcal{D}$ and the stochastic policy $\pi_{\theta_{\pi}}$. However, in the SAC-Lagrangian algorithm, we consider (\ref{SAC_Lagrangian}) as the objective instead of (\ref{ent_objective}). Thus, 
the parameter vector of the actor DNN $\theta_{\pi}$ is updated by decreasing the following actor loss function.
\begin{eqnarray}
J_{a}(\theta_{\pi})&=&E_{z\sim\mathcal{D},a\sim\pi_{\theta_{\pi}}}[\alpha\log(\pi_{\theta_{\pi}}(a|z))\nonumber\\
&& \ \ \ -(Q_{\theta_{r}}(z,a)+\kappa Q_{\theta_{s}}(z,a))].\label{SAC_actor_update}
\end{eqnarray}
The Lagrange multiplier $\kappa$ is updated by decreasing the following loss function.
\begin{eqnarray}
J_{L}(\kappa)=E_{z_0\sim p_{0}^{z},a\sim\pi_{\theta_{\pi}}}\left[\kappa(Q_{\theta_{s}}(z_0,a)-l_{STL})\right],\label{SAC_kappa_update}
\end{eqnarray}
where $E_{z_0\sim p_{0}^{z},a\sim\pi_{\theta_{\pi}}}[\cdot]$ is the expected value with respect to $p_{0}^{z}$ and $\pi_{\theta_{\pi}}$. The entropy temperature $\alpha$ is updated by decreasing the following loss function.
\begin{eqnarray}
J_{temp}(\alpha)=E_{z\sim \mathcal{D},a\sim\pi_{\theta_{\pi}}}\left[\alpha(-\log(\pi_{\theta_{\pi}}(a|z))-\mathcal{H}_{0})\right],\label{alpha_update}
\end{eqnarray}
where $\mathcal{H}_{0}$ is a lower bound which is a hyper-parameter. In \cite{SAC}, the parameter $\mathcal{H}_0$ is selected based on the dimensionality of the action space. Additionally, in the SAC algorithm, to mitigate the positive bias in updates of $\theta_{\pi}$, the double Q-learning technique \cite{TD3} is adopted, where we prepare two critic DNNs and two target critic DNNs. Thus, in the SAC-Lagrangian, we also adopt the technique.

\subsection{Pre-training and fine-tuning}
In this study, it is important to satisfy the given STL constraint. In order to learn a policy satisfying a given STL formula, the agent needs many experiences satisfying the formula. However, it is difficult to collect the experiences considering both the control performance index and the STL constraint in the early learning stage since the agent may prioritize to optimize its policy with respect to the control performance index. Thus, we propose a two-phase learning algorithm. In the first phase, which is called \textit{pre-train}, the agent focuses on learning a policy satisfying a given STL formula $\Phi$ to store experiences receiving high STL rewards to a replay buffer $\mathcal{D}$, that is, the agent learns its policy considering only STL-rewards.    

\noindent \textbf{Pre-training for DDPG-Lagrangian}

The parameter vector of the actor DNN $\theta_{\mu}$ is updated by decreasing 
\begin{eqnarray}
J_{a}(\theta_{\mu})=E_{z\sim\mathcal{D}}\left[-Q_{\theta_{s}}(z,\mu_{\theta_{\mu}}(z))\right]\label{Pretrain_DDPG_update}
\end{eqnarray}
instead of (\ref{DDPG_actor_update}). On the other hand, $\theta_{s}$ is updated by (\ref{STL_DDPG_reward_critic_update}).

\noindent \textbf{Pre-training for SAC-Lagrangian}

The parameter vector of the actor DNN $\theta_{\pi}$ is updated by decreasing
\begin{eqnarray}
J_{a}(\theta_{\pi})=E_{z\sim\mathcal{D},a\sim\pi_{\theta_{\pi}}}\left[\alpha\log(\pi_{\theta_{\pi}}(a|z))-Q_{\theta_{s}}(z,a)\right]\label{Pretrain_update}
\end{eqnarray}
instead of (\ref{SAC_actor_update}). On the other hand, $\theta_{s}$ is updated by (\ref{STL_reward_critic_update}), where $V_{\theta_{s}}^{-}$ is computed by
\begin{eqnarray}
V_{\theta_{s}^{-}}(z')=E_{a'\sim\pi_{\theta_{\pi}}}[Q_{\theta_{s}^{-}}(z',a')-\alpha\log(\pi_{\theta_{\pi}}(a'|z'))].\nonumber
\end{eqnarray}
In the second phase, which is called \textit{fine-tune}, the agent learns the optimal policy constrained by the given STL formula. In the DDPG-Lagrangian algorithm, the actor DNN $\theta_{\mu}$ is updated by (\ref{DDPG_actor_update}). In the SAC-Lagrangian algorithm, the actor DNN $\theta_{\pi}$ is updated by (\ref{SAC_actor_update}). 

\noindent \textbf{Remark:} The two-phase learning may become unstable temporally because it discontinuously changes the objective functions. In such a case, we may start the second phase with changing the objective functions from those used in the first phase smoothly and slowly. 

\subsection{Pre-process}
\begin{figure}[h]
\begin{center}
  \includegraphics[width=8.5cm]{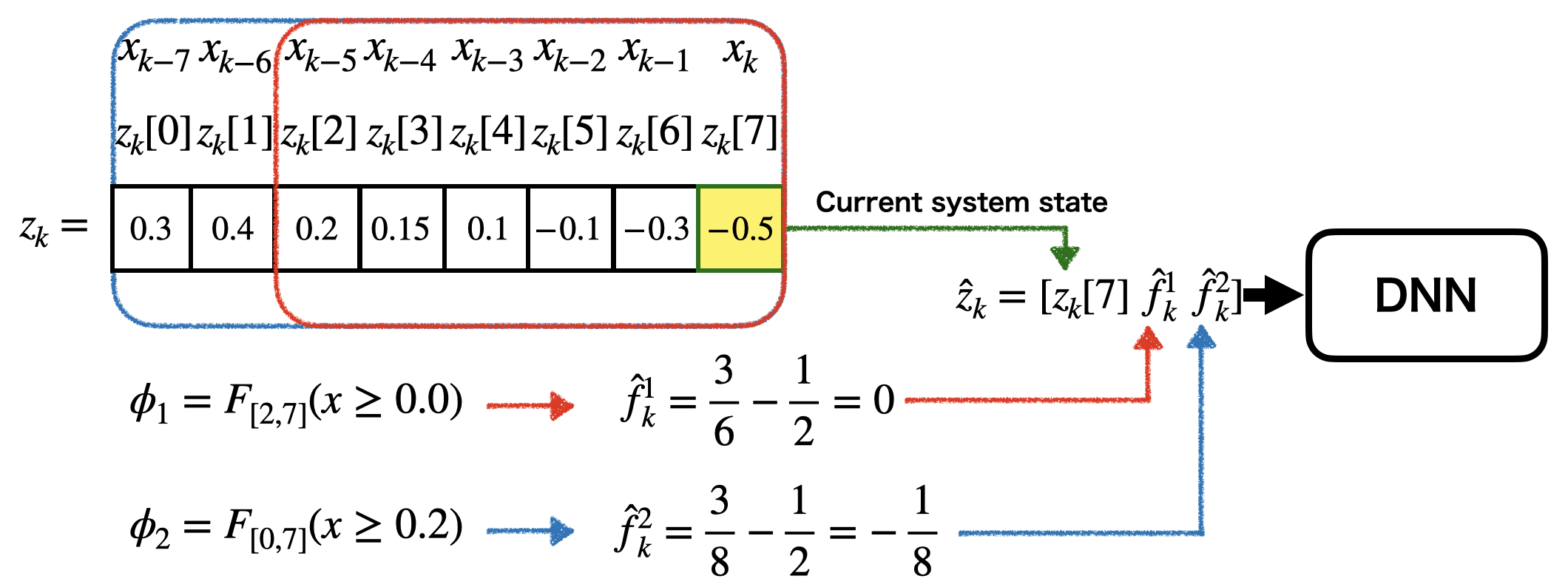}
  \caption{Example of constructing a pre-processed state. We consider the 1-dimensional system and the two STL sub-formulae: $\phi_{1}=F_{[2,7]}(x\ge0.0)$ and $\phi_{2}=F_{[0,7]}(x\ge0.2)$. For each sub-formula, we compute the flag value using the extended state $z_k$, which is regarded as a continuous value in $[-0.5,0.5]$. After that, we construct the pre-processed state using $z_{k}[\tau-1](=x_k)$, $\hat{f}_{k}^{1}$, and $\hat{f}_{k}^{2}$ and input it to DNNs.}
  \label{pre_processing}
\end{center}
\end{figure}

If $\tau$ is a large value, it is difficult for the agent to learn its policy due to the large dimensionality of the extended state space. Then, \textit{pre-process} is useful in order to reduce the dimensionality, which is related to \cite{Venkataraman}. In the previous study, a flag state for each sub-formula is defined as a discrete state. The flag discrete state space is combined with the system's discrete state space. On the other hand, in this study, it is assumed that the system state space is continuous. If we use the discrete flag states, the pre-processed state space is a hybrid state space that has discrete values and continuous values. Thus, we consider the flag state as a continuous value and input it to DNNs as shown in Fig.\ \ref{pre_processing}.  

We introduce a flag value $f^{i}$ for each STL sub-formula $\phi_{i}$, where it is assumed that $k_e^{i}=\tau-1,\ \forall i\in\{1,2,...,M\}$.

\noindent \textbf{Definition 3 (Pre-process):} For an extended state $z$, a flag value $f^{i}$ of an STL sub-formula $\phi_{i}$ is defined as follows:

\noindent (i) For $\phi_{i}=G_{[k_{s}^{i},\tau-1]}\varphi_{i}$,
\begin{eqnarray}
f^{i}=\max\left\{\frac{\tau-l}{\tau-k_{s}^{i}}\ \middle|\ l\in\{k_{s}^{i},...,\tau-1\}\hspace{0.5cm} \right.\nonumber\\ 
\biggl. \land(\forall l'\in\{l,...,\tau-1\},\ z[l']\models\varphi_{i}) \biggr\}.\label{G}
\end{eqnarray}

\noindent (ii) For $\phi_{i}=F_{[k_{s}^{i},\tau-1]}\varphi_{i}$,
\begin{eqnarray}
f^{i}=\max\left\{\frac{l-k_{s}^{i}+1}{\tau-k_{s}^{i}}\ \middle|\hspace{3cm} \right. \nonumber\\
\biggl.\hspace{2cm} l\in\{k_{s}^{i},...,\tau-1\}\land z[l]\models\varphi_{i}\biggr\}.
\label{F}
\end{eqnarray}
Note that $\max\emptyset=-\infty$ and the flag value represents the normalized time lying in $(0,1]\cap\{-\infty\}$. Intuitively, for $\phi_{i}=G_{[k_{s}^{i},\tau-1]}\varphi_{i}$, the flag value indicates the time duration in which $\phi_{i}$ is always satisfied, whereas, for $\phi_{i}=F_{[k_{s}^{i},\tau-1]}\varphi_{i}$, the flag value indicates the instant when $\varphi_{i}$ is satisfied. The flag values $f^{i},\ i\in\{1,2,...,M\}$ calculated by (\ref{G}) or (\ref{F}) are transformed into $\hat{f}^{i}$ as follows:
\begin{eqnarray}
\hat{f}^{i}=\begin{cases}
	f^{i}-\frac{1}{2} & \text{if }f^{i}\neq-\infty,\\
	-\frac{1}{2} & \text{otherwise}.
\end{cases}\label{preprocess_flag}
\end{eqnarray}
The transformed flag values $\hat{f}^{i}$ are used as inputs to DNNs to prevent positive biases of the flag values and inputting $-\infty$ to DNNs. We compute the flag value for each STL sub-formula and construct a flag state $\hat{f}=[\hat{f}^{1}\ \hat{f}^{2}\ ...\ \hat{f}^{M}]^{\top}$, which is called pre-processing. We use the pre-processed state $\hat{z}=[z[\tau-1]^{\top}\ \hat{f}^{\top}]^{\top}$ as an input to DNNs instead of the extended state $z$. 

\noindent \textbf{Remark:} It is important to ensure the Markov property of the pre-processed state for the agent to learn its policy. If $k_e^{i}=\tau-1,\ \forall i\in\{1,2,...,M\}$, then the pre-processed state $\hat{z}$ satisfies the Markov property. We consider the current pre-processed state $\hat{z}=[z[\tau-1]^{\top}\ \hat{f}^{\top}]^{\top}$ and the next pre-processed state $\hat{z}'=[z'[\tau-1]^{\top}\ \hat{f}']^{\top}$. $z'[\tau-1]$ is generated by $p_{f}(\cdot|z[\tau-1],a)$, where $a$ is the current action. Therefore, $z'[\tau-1]$ depends on $z[\tau-1]$ and the current action $a$. For each transformed flag value $\hat{f}^{i},\ i\in\{1,2,...,M\}$, it is updated by
\begin{enumerate}
\item $\phi_i=G_{[k_s^{i},\tau-1]}\varphi_i$
\begin{eqnarray}
\hat{f}^{i'}=\begin{cases}
	\min\left\{\hat{f}^{i}+\frac{1}{\tau-k_s^{i}},\frac{1}{2}\right\}, & x'\models \varphi_i,\\
	-\frac{1}{2}, & x'\not\models \varphi_i,
\end{cases}
\end{eqnarray}
\item $\phi_i=F_{[k_s^i,\tau-1]}\varphi_i$
\begin{eqnarray}
\hat{f}^{i'}=\begin{cases}
	\frac{1}{2}, & x'\models \varphi_i,\\
	\max\left\{\hat{f}^{i}-\frac{1}{\tau-k_s^i},-\frac{1}{2}\right\}, & x'\not\models \varphi_i,
\end{cases}
\end{eqnarray}
\end{enumerate}
where $x'\sim p_{f}(\cdot|z[\tau-1],a)$. The transformed flag values are updated by the next system's  state. Therefore, the next transformed flag values $\hat{f}^{i'},\ i\in\{1,2,...,M\}$ depends on $\hat{f}^{i}$, $z[\tau-1]$, and the current action $a$. Thus, the Markov property of the pre-processed state holds. 

\begin{figure}[h]
\begin{center}
  \includegraphics[width=8.5cm]{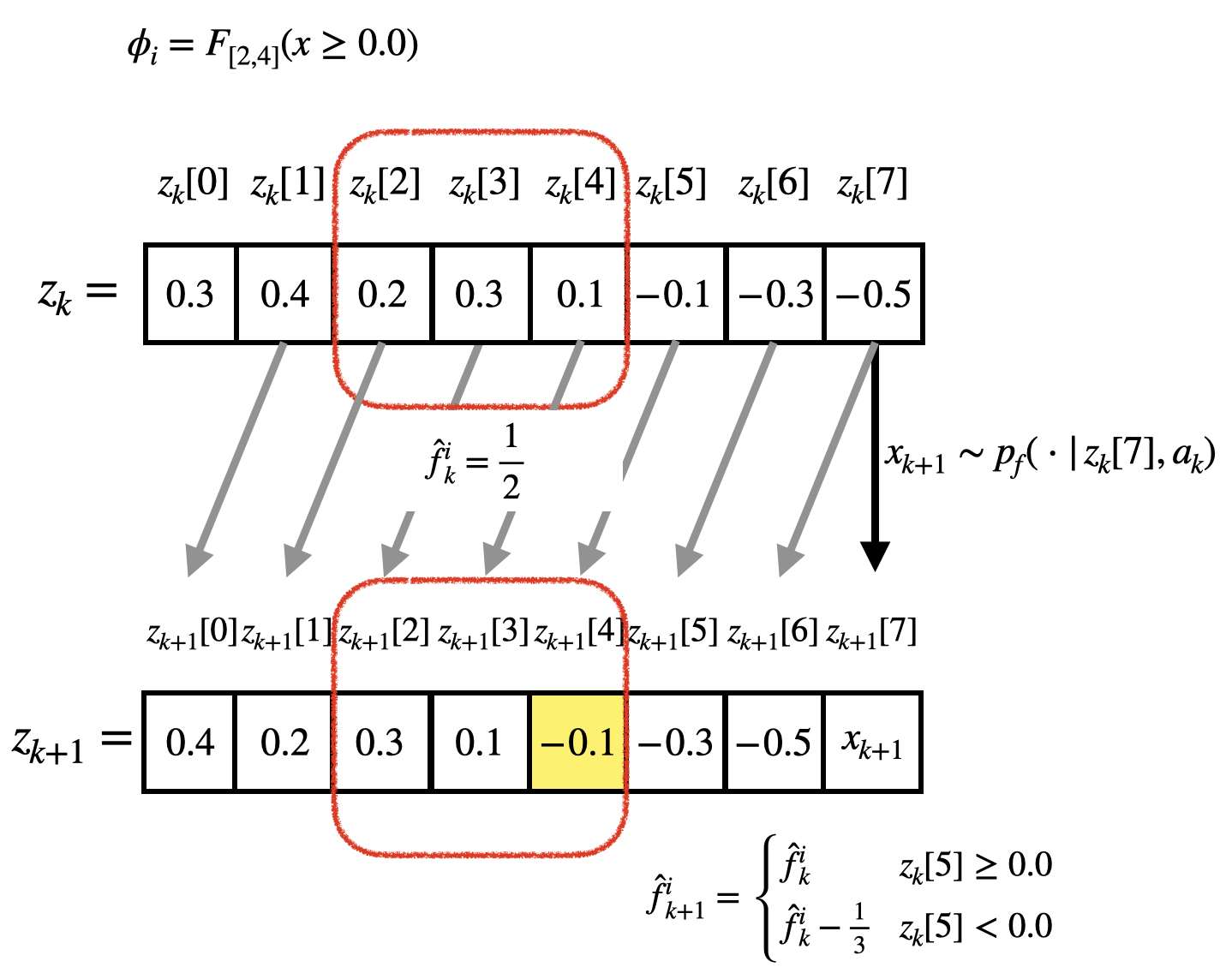}
  \caption{Example of a sub-formula $\phi_{i}$ with $k_{e}^{\max}\ge k_{e}^{i}+1$. We consider the 1-dimensional system and  $k_{e}^{\max}=7$ ($\tau=8$). For the sub-formula $\phi_i=F_{[2,4]}(x\ge0.0)$, $z_{k+1}[7](=x_{k+1})$ depends on $z_{k}[7](=x_k)$ and $a_k$. However, $\hat{f}_{k+1}^{i}$ depends on $\hat{f}_{k}^{i}$ and $z_{k}[5]$. If the pre-processed state is given by $[z_{k}[7]\ \hat{f}_{k}]^{\top}$, the agent with DNNs observes the environment partially. Then, the agent also needs $z_{k}[5]$ and $z_{k}[6]$ as parts of the pre-processed state.}
  \label{general_case_a}
\end{center}
\end{figure}
\begin{figure}[h]
\begin{center}
  \includegraphics[width=8.5cm]{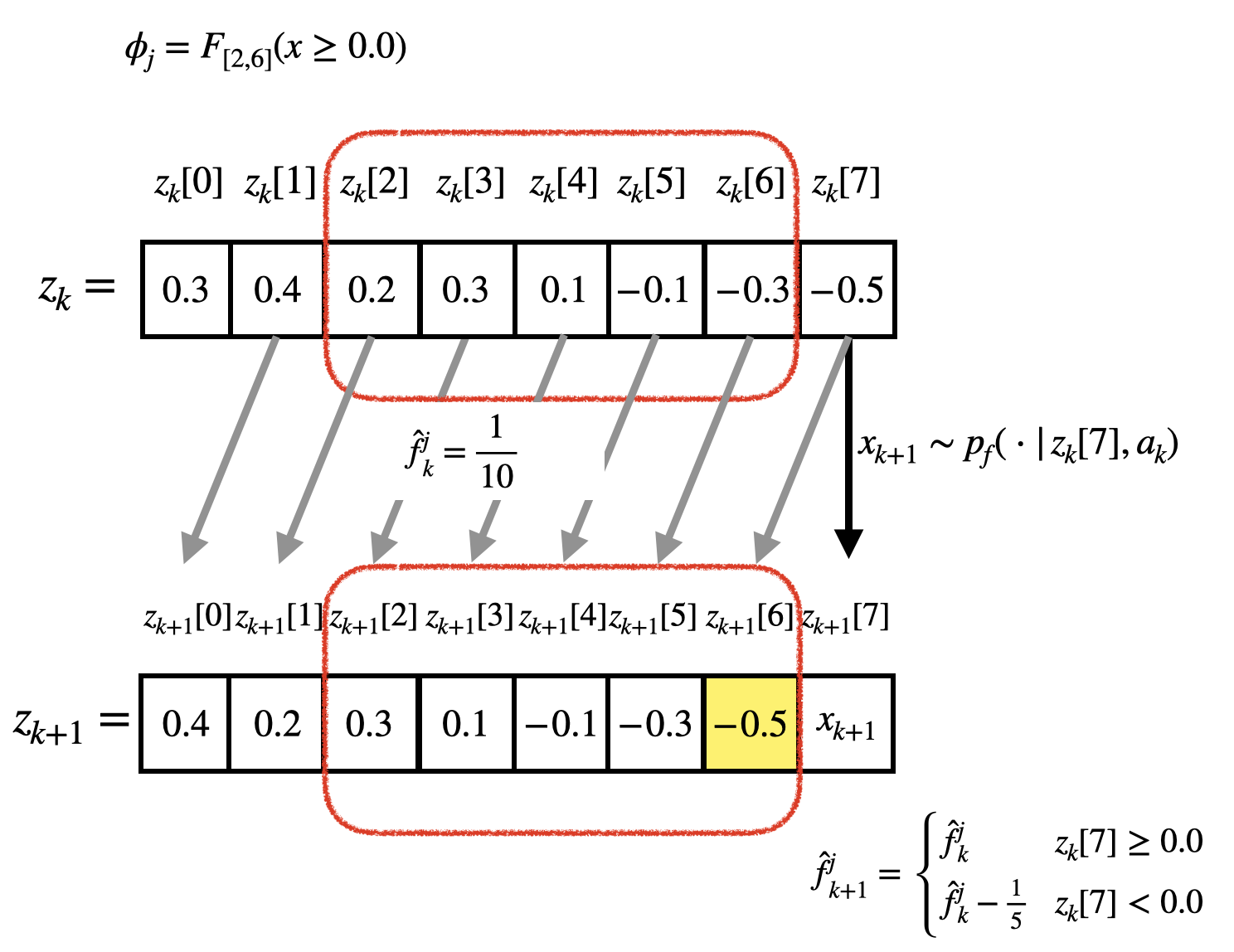}
  \caption{Example of the sub-formula $\phi_j$ with $k_{e}^{\max}= k_{e}^{j}+1$. We consider the 1-dimensional system and  $k_{e}^{\max}=7$ ($\tau=8$). For $\phi_{j}=F_{[2,6]}(x\ge0.0)$, that is $k_{e}^{\max}-k_{e}^{j}=1$, the transformed flag value can be updated by $[z_{k}[7]\ \hat{f}_{k}]^{\top}$ only.}
  \label{general_case_b}
\end{center}
\end{figure}

On the other hand, in the case where $k_{e}^{\max}\ge k_{e}^{\min}+1$, we must include $z[\tau-k_{e}^{\max}+k_{e}^{\min}],...,z[\tau-1]$ to the pre-processed state $\hat{z}$ in order to ensure the Markov property, where $k_e^{\max}=\max_{i\in\{1,2,...,M\}}k_e^i$ and $k_e^{\min}=\min_{i\in\{1,2,...,M\}}k_e^i$. For example, as shown in Fig.\ \ref{general_case_a}, there may be some transformed flag values that are updated with information other than $[z[\tau-1]^{\top}\ \hat{f}]^{\top}$ and the current action. Note that, in the case $k_{e}^{\max}=k_{e}^{j}+1$ as shown in Fig.\ \ref{general_case_b}, the transformed flag value $\hat{f}^{j}$ is updated by $[z[\tau-1]^{\top}\ \hat{f}]^{\top}$, that is, the agent with DNNs can learn its policy using $[z[\tau-1]^{\top}\ \hat{f}]^{\top}$ when $k_{e}^{\max}=k_{e}^{\min}+1$. As the difference $k_{e}^{\max}-k_{e}^{\min}$ increases, we need to include more past system states in the pre-processed state. 

For simplicity, in this study, we focus on the case where $k_e^{i}=\tau-1,\ \forall i\in\{1,2,...,M\}$. Then, the pre-processing is most effective in terms of reducing the dimensionality of the extended state space. 

\subsection{Algorithm}
Our proposed algorithm to design an optimal policy under the given STL constraint is presented in \textbf{Algorithm \ref{Lag_Pretrain}}. In line 1, we select a DRL algorithm such as the DDPG algorithm and the SAC algorithm. From line 2 to 4, we initialize the parameter vectors of the DNNs, the entropy temperature (if the algorithm is the SAC-Lagrangian algorithm), and the Lagrange multiplier. In line 5, we initialize a replay buffer $\mathcal{D}$. In line 6, we set the number of the repetition of pre-training $K_{pre}$. In line 7, we initialize a counter for updates. In line 9, the agent receives an initial state $x_0\sim p_0$. From line 10 to 11, the agent sets the initial extended state $z_{0}=[x_{0}^{\top}\ ...\ x_{0}^{\top}]^{\top}$ and computes the pre-processed state $\hat{z}_0$. One learning step is done between line 13 and 25. In line 13, the agent determines an action $a_{k}$ based on the pre-processed state $\hat{z}_{k}$ for an exploration. In line 14, the state of the system changes by the determined action $a_{k}$ and the agent receives the next state $x_{k+1}$, the reward $r_{k}$, and the STL-reward $s_{k}$. From line 15 to 16, the agent sets the next extended state $z_{k+1}$ using $x_{k+1}$ and $z_{k}$ and computes the next pre-processed state $\hat{z}_{k+1}$. In line 17, the agent stores the experience $(\hat{z}_{k},a_{k},\hat{z}_{k+1},r_{k},s_{k})$ in the replay buffer $\mathcal{D}$. In line 18, the agent samples $I$ experiences $\{(\hat{z}^{(i)},a^{(i)},\hat{z}'^{(i)},r^{(i)},s^{(i)})\}_{i=1}^{I}$ from the replay buffer $\mathcal{D}$ randomly. If the learning counter is $c< K_{pre}$, the agent pre-trains the parameter vectors in \textbf{Algorithm 3}. Then, the parameter vectors of the reward critic DNN $\theta_{r}$ and the STL-reward critic DNN $\theta_{s}$ are updated by (\ref{reward_DDPG_critic_update}) and (\ref{STL_DDPG_reward_critic_update}) (or  (\ref{reward_critic_update}) and (\ref{STL_reward_critic_update})), respectively. The parameter vector of the actor DNN $\theta_{\mu}$ (or $\theta_{\pi}$) is updated by (\ref{Pretrain_DDPG_update}) (or (\ref{Pretrain_update})). In the SAC-based algorithm, the entropy temperature $\alpha$ is updated by (\ref{alpha_update}). On the other hand, if the learning counter is $c\ge K_{pre}$, the agent fine-tunes the parameter vectors in \textbf{Algorithm 4}. Then, the parameter vector of the actor DNN $\theta_{\mu}$ (or $\theta_{\pi}$) is updated by (\ref{DDPG_actor_update}) (or (\ref{SAC_actor_update})) and the other parameter vectors are updated same as the case $c< K_{pre}$. The Lagrange multiplier is updated by (\ref{DDPG_kappa_update}) (or (\ref{SAC_kappa_update})). In line 24, the agent updates the parameter vectors of the target DNNs by (\ref{soft_update}). In line 25, the learning counter is updated. The agent repeats the process between lines 13 and 25 in a learning episode.  
\begin{algorithm}[h]
\caption{Two-phase DRL-Lagrangian to design an optimal policy under an STL constraint.}         
\label{Lag_Pretrain}                          
\begin{algorithmic}[1]
\STATE Select a DRL algorithm such as DDPG and SAC.             
\STATE Initialize parameter vectors of main DNNs. 
\STATE Initialize parameter vectors of target DNNs.
\STATE Initialize an entropy temperature and a Lagrange multiplier $\alpha,\ \kappa$.
\STATE Initialize a replay buffer $\mathcal{D}$.
\STATE Set the number of the repetition of pre-training $K_{pre}$.
\STATE Initialize learning counter $c\leftarrow0$.
\FOR{$\text{Episode}=1,...,\text{MAX EPISODE}$}
\STATE Receive an initial state $x_{0}\sim p_{0}$.
\STATE Set the initial extended state $z_{0}$ using $x_{0}$.
\STATE Compute the pre-processed state $\hat{z}_{0}$ by \textbf{Algorithm 2}.
\FOR{Discrete-time step $k=0,...,K$}
\STATE Determine an action $a_{k}$ based on the state $\hat{z}_{k}$.
\STATE Execute $a_{k}$ and receive the next state $x_{k+1}$ and the reward $r_{k}$ and the STL-reward $s_{k}$.
\STATE Set the next extended state $z_{k+1}$ using $x_{k+1}$ and $z_{k}$.
\STATE Compute the next pre-processed state $\hat{z}_{k+1}$ by \textbf{Algorithm 2}.
\STATE Store the experience $(\hat{z}_{k},a_{k},\hat{z}_{k+1},r_{k},s_{k})$ in the replay buffer $\mathcal{D}$.
\STATE Sample $I$ experiences \\$\{(\hat{z}^{(i)},a^{(i)},\hat{z}'^{(i)},r^{(i)},s^{(i)}) \}_{i=1,...,I}$\\ from $\mathcal{D}$ randomly.
\IF{$c< K_{pre}$}
\STATE Pre-training by \textbf{Algorithm 3}. 
\ELSE
\STATE Fine-tuning by \textbf{Algorithm 4}. 
\ENDIF
\STATE Update the target DNNs by (\ref{soft_update}).
\STATE $c\leftarrow c + 1$.
\ENDFOR
\ENDFOR
\end{algorithmic}
\end{algorithm}

\begin{algorithm}[h]               
\caption{Pre-processing of the extended state}         
\label{preprocess}                          
\begin{algorithmic}[1]
\STATE \textbf{Input:} The extended state $z$ and the STL sub-formulae $\{\phi_{i}\}_{i=1}^{M}$.    
\FOR{$i=1,...,M$}   
\IF{$\phi_{i}=G_{[k_{s}^{i},\tau-1]}\varphi_{i}$}
\STATE Compute the flag value $f^{i}$ by (\ref{G}).
\ENDIF
\IF{$\phi_{i}=F_{[k_{s}^{i},\tau-1]}\varphi_{i}$}
\STATE Compute the flag value $f^{i}$ by (\ref{F}).
\ENDIF       
\ENDFOR
\STATE Set the flag state $\hat{f}=[\hat{f}^{1}\ \hat{f}^{2}\ ...\ \hat{f}^{M}]^{\top}$.
\STATE \textbf{Output:} The pre-processed state $\hat{z}=[z[\tau-1]^{\top}\ \hat{f}^{\top}]^{\top}$.    
\end{algorithmic}
\end{algorithm}

\begin{algorithm}[h]               
\caption{Pre-training}         
\label{pretraining}                          
\begin{algorithmic}[1]
\STATE \textbf{Input:} \\The experiences $\{(\hat{z}^{(i)},a^{(i)},\hat{z}'^{(i)},r^{(i)},s^{(i)})\}_{i=1,2,...,I}$ and parameters $\theta_{\pi},\ \theta_{r},\ \theta_{s},\ \alpha$.   
\STATE The parameter vector $\theta_{r}$ is updated by (\ref{reward_DDPG_critic_update}) or (\ref{reward_critic_update}).   
\STATE The parameter vector $\theta_{s}$ is updated by (\ref{STL_DDPG_reward_critic_update}) or (\ref{STL_reward_critic_update}) .   
\STATE The parameter vector $\theta_{\pi}$ is updated by (\ref{Pretrain_DDPG_update}) or (\ref{Pretrain_update}).  
\IF{SAC-based algorithm} 
\STATE The entropy temperature $\alpha$ is updated by (\ref{alpha_update}).   
\ENDIF
\STATE \textbf{Output: } $\theta_{\pi},\ \theta_{r},\ \theta_{s},\ \alpha$
\end{algorithmic}
\end{algorithm}

\begin{algorithm}[h!]               
\caption{Fine-tuning}         
\label{finetuning}                          
\begin{algorithmic}[1]
\STATE \textbf{Input:} \\The experiences $\{(\hat{z}^{(i)},a^{(i)},\hat{z}'^{(i)},r^{(i)},s^{(i)})\}_{i=1,2,...,I}$ and parameters $\theta_{\pi},\ \theta_{r},\ \theta_{s},\ \alpha,\ \kappa$.    
\STATE The parameter vector $\theta_{r}$ is updated by (\ref{reward_DDPG_critic_update}) or (\ref{reward_critic_update}).   
\STATE The parameter vector $\theta_{s}$ is updated by (\ref{STL_DDPG_reward_critic_update}) or (\ref{STL_reward_critic_update}).   
\STATE The parameter vector $\theta_{\pi}$ is updated by (\ref{DDPG_actor_update}) or (\ref{SAC_actor_update}).  
\IF{SAC-based algorithm} 
\STATE The entropy temperature $\alpha$ is updated by (\ref{alpha_update}). 
\ENDIF  
\STATE The Lagrange multiplier $\kappa$ is updated by (\ref{DDPG_kappa_update}) or (\ref{SAC_kappa_update}).
\STATE \textbf{Output: } $\theta_{\pi},\ \theta_{r},\ \theta_{s},\ \alpha,\ \kappa$
\end{algorithmic}
\end{algorithm}

\section{Example}
\begin{figure}[h]
\begin{center}
  \includegraphics[width=7.0cm]{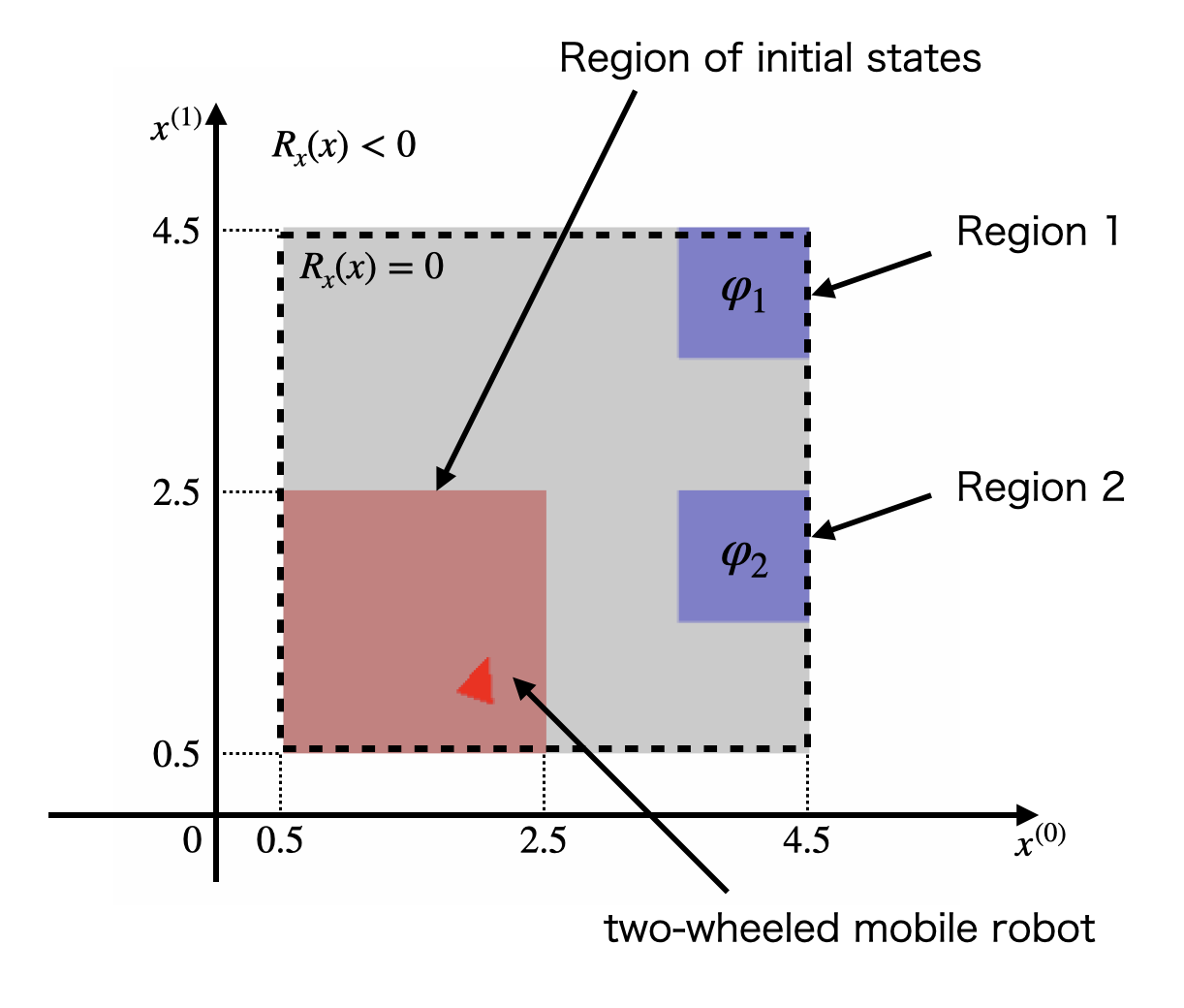}
  \caption{Control of a two-wheeled mobile robot under an STL constraint. The working area is $0.5\le x^{(0)}\le4.5,\ 0.5\le x^{(1)}\le4.5$ colored gray. The initial state of the system is sampled randomly in $0.5\le x^{(0)}\le 2.5,\ 0.5\le x^{(1)}\le 2.5,\ -\pi/2\le x^{(2)}\le \pi/2$ colored red. The region 1 labeled by $\varphi_{1}$ is $3.5\le x^{(0)}\le 4.5,\ 3.5\le x^{(1)}\le 4.5$ and the region 2 labeled by $\varphi_2$ is $3.5\le x^{(0)}\le 4.5,\ 1.5\le x^{(1)}\le 2.5$. These regions are colored blue.}
  \label{Example_pic}
\end{center}
\end{figure}

We consider STL-constrained optimal control problems for a two-wheeled mobile robot shown in Fig.\ \ref{Example_pic}, where its working area $\Omega$ is $\{(x^{(0)},x^{(1)})|\ 0.5\le x^{(0)}\le 4.5,\ 0.5\le x^{(1)}\le 4.5\}$. Let $x^{(2)}$ be the steering angle with $x^{(2)}\in[-\pi,\pi]$. A discrete-time model of the robot is described by
\begin{eqnarray}
\begin{bmatrix}
	x_{k+1}^{(0)}\\
	x_{k+1}^{(1)}\\
	x_{k+1}^{(2)}	
\end{bmatrix}=\begin{bmatrix}
	x_{k}^{(0)} + \Delta a_{k}^{(0)}\cos(x_{k}^{(2)})\\
	x_{k}^{(1)} + \Delta a_{k}^{(0)}\sin(x_{k}^{(2)})\\
	x_{k}^{(2)} + \Delta a_{k}^{(1)}
\end{bmatrix}+\Delta_{w}\begin{bmatrix}
	w_{k}^{(0)}\\
	w_{k}^{(1)}\\
	w_{k}^{(2)}
\end{bmatrix},\label{example}
\end{eqnarray}
where $x_{k}=[x_{k}^{(0)}\ x_{k}^{(1)}\ x_{k}^{(2)}]^{\top}\in\mathbb{R}^{3}$, $a_{k}=[a_{k}^{(0)}\ a_{k}^{(1)}]^{\top}\in[-1,1]^{2}$, and $w_{k}=[w_{k}^{(0)}\ w_{k}^{(1)}\ w_{k}^{(2)}]^{\top}\in\mathbb{R}^{3}$. $w_{k}^{(i)},\ i\in\{0,1,2\}$ is sampled from a standard normal distribution $\mathcal{N}(0,1)$. We assume that $\Delta=0.1$ and $\Delta_{w}=0.01I$, where $I$ is the unit matrix. The initial state of the system is sampled randomly in $0.5\le x^{(0)}\le 2.5,\ 0.5\le x^{(1)}\le 2.5,\ -\pi/2\le x^{(2)}\le \pi/2$. The region 1 is $\{(x^{(0)},x^{(1)})|\ 3.5\le x^{(0)}\le 4.5,\ 3.5\le x^{(1)}\le 4.5\}$ and the region 2 is $\{(x^{(0)},x^{(1)})|\ 3.5\le x^{(0)}\le 4.5,\ 1.5\le x^{(1)}\le 2.5\}$. We consider the following two constraints. 

\noindent \textbf{Constraint 1 (Recurrence): }
 At any time in the time interval $[0,900]$, the robot visits both the regions 1 and 2 before $99$ time steps are elapsed, where there is no constraint for the order of the visits.

\noindent \textbf{Constraint 2 (Stabilization): }
The robot visits the region 1 or 2 in the time interval $[0,450]$ and stays there for $49$ time steps.

These constraints are described by the following STL formulae. 

\noindent \textbf{Formula 1:}
\begin{eqnarray}
\Phi_{1} &=& G_{[0,900]}(F_{[0,99]}\varphi_{1}\land F_{[0,99]}\varphi_{2}),\label{specification_1}
\end{eqnarray}
\noindent \textbf{Formula 2:}
\begin{eqnarray}
\Phi_{2} &=& F_{[0,450]}(G_{[0,49]}\varphi_{1}\lor G_{[0,49]}\varphi_{2}),\label{specification_2}
\end{eqnarray}
where  
\begin{eqnarray}
\varphi_{1} &=& ((3.5\le x^{(0)}\le 4.5)\land(3.5\le x^{(1)}\le 4.5)),\nonumber\\
\varphi_{2} &=& ((3.5\le x^{(0)}\le 4.5)\land(1.5\le x^{(1)}\le 2.5)).\nonumber
\end{eqnarray}
We consider the following reward function 
\begin{eqnarray}
R_{z}(z,a)=R_{x}(z[\tau-1])+R_{a}(a),\label{reward_func}
\end{eqnarray}
where
\begin{eqnarray}
R_{x}(x)&=&\min\{x^{(0)}-0.5,4.5-x^{(0)},\nonumber\\
&&\ \ \ \ \ x^{(1)}-0.5,\ 4.5-x^{(1)},\ 0.0\},\label{working_area_reward}\\
R_{a}(a)&=&-||a||_{2}^{2}.\label{fuel_cost}
\end{eqnarray}
(\ref{working_area_reward}) is the term for keeping the working area. As the agent moves away from the working area, the agent receives a larger negative reward. (\ref{fuel_cost}) is the term for fuel costs.
 
\subsection{Evaluation}
We apply the SAC-Lagrangian algorithm to design a policy constrained by an STL formula. In all simulations, the DNNs had two hidden layers, all of which have 256 units, and all layers are fully connected. The activation functions for the hidden layers and the outputs of the actor DNN are the rectified linear unit functions and hyperbolic tangent functions, respectively. We normalize $x^{(0)}$ and $x^{(1)}$ as $x^{(0)}-2.5$ and $x^{(1)}-2.5$, respectively. The size of the replay buffer $\mathcal{D}$ is $1.0\times10^{5}$, and the size of the mini-batch is $I=64$. We use \textit{Adam} \cite{Adam} as the optimizers for all main DNNs, the entropy temperature, and the Lagrange multiplier. The learning rate of the optimizer for the Lagrange multiplier is $1.0\times10^{-5}$ and the learning rates of the other optimizers are $3.0\times10^{-4}$. The soft update rate of the target network is $\xi=0.01$. The discount factor is $\gamma=0.99$. The target for updating the entropy temperature $\mathcal{H}_{0}$ is $-2.0$. The STL-reward parameter is $\beta=100$. The agent learns its control policy for $6.0\times10^{5}$ steps. The initial parameters of both the entropy temperature and the Lagrange multiplier are $1.0$. For performance evaluation, we introduce the following three indices:
\begin{itemize}
\item  a \textbf{reward learning curve} shows the mean of the sum of rewards $\sum_{k=0}^{K}\gamma^{k}R_{z}(z_{k},a_{k})$ for 100 trajectories,
\item  an \textbf{STL-reward learning curve} shows the mean of the sum of STL-rewards $\sum_{k=0}^{K}\gamma^{k}R_{STL}(z_{k})$ for 100 trajectories, and
\item a \textbf{success rate} shows the number of trajectories satisfying the given STL constraint for 100 trajectories.
\end{itemize}
We prepare $100$ initial states sampled from $p_0$ and generate $100$ trajectories using the learned policy for each evaluation. We show the results for $K_{pre}=0$ (\textbf{Case 1}) and $K_{pre}=300000$ (\textbf{Case 2}). We do not use pre-training in \textbf{Case 1}. All simulations were done on a computer with AMD Ryzen 9 3950X 16-core processor, NVIDIA (R) GeForce RTX 2070 super, and 32GB of memory and were conducted using the Python software.

\subsubsection{Formula 1}
\begin{figure}[h]
\begin{center}
  \includegraphics[width=8.0cm]{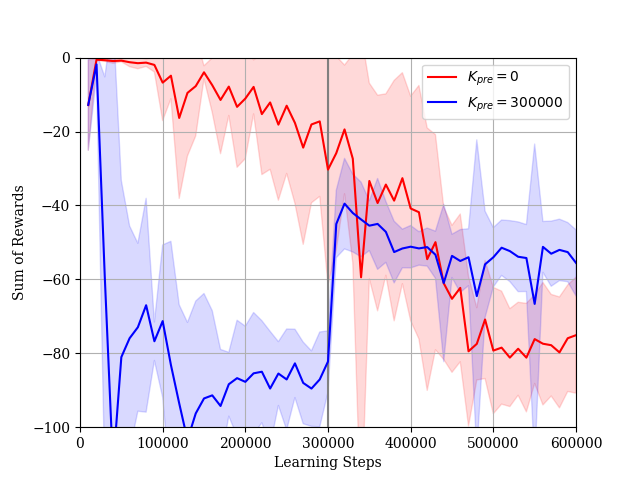}
  \caption{Reward learning curves for the formula $\Phi_1$. The red and blue curves show the results of $K_{pre}=0$ (\textbf{Case 1}) and $K_{pre}=300000$ (\textbf{Case 2}), respectively. The solid curves and the shades represent the average results and standard deviations over $10$ trials with different random seeds, respectively. The gray line shows $300000$ steps. }
  \label{Phi_1_rewards}
\end{center}
\end{figure}

\begin{figure}[h]
\begin{center}
  \includegraphics[width=8.0cm]{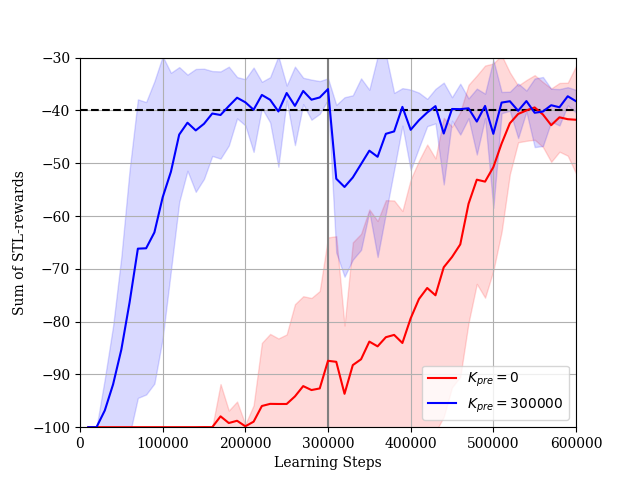}
  \caption{STL-reward learning curves for the formula $\Phi_1$. The red and blue curves show the results of $K_{pre}=0$ (\textbf{Case 1}) and $K_{pre}=300000$ (\textbf{Case 2}), respectively. The solid curves and the shades represent the average results and standard deviations over $10$ trials with different random seeds, respectively. The dashed line shows the threshold $l_{STL}=-40$. The gray line shows $300000$ steps. }
  \label{Phi_1_stlrewards}
\end{center}
\end{figure}

\begin{figure}[h]
\begin{center}
  \includegraphics[width=8.0cm]{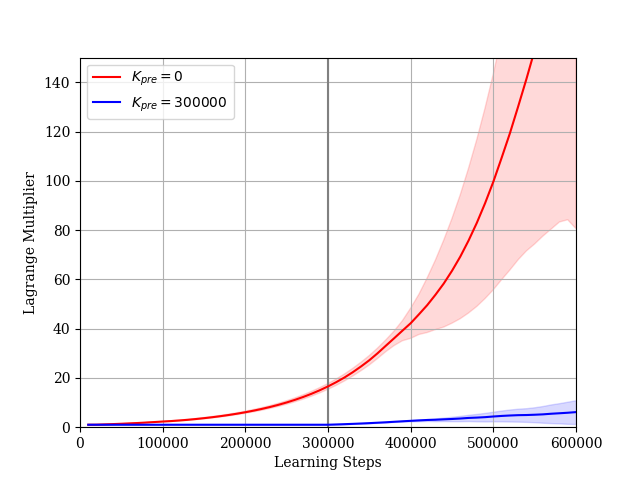}
  \caption{Curves of Lagrange multiplier $\kappa$ for the formula $\Phi_1$. The red and blue curves show the results of $K_{pre}=0$ (\textbf{Case 1}) and $K_{pre}=300000$ (\textbf{Case 2}), respectively. The solid curves and the shades represent the average results and standard deviations over $10$ trials with different random seeds, respectively. The gray line shows $300000$ steps.}
  \label{Phi_1_kappa}
\end{center}
\end{figure}

\begin{figure}[h]
\begin{center}
  \includegraphics[width=8.0cm]{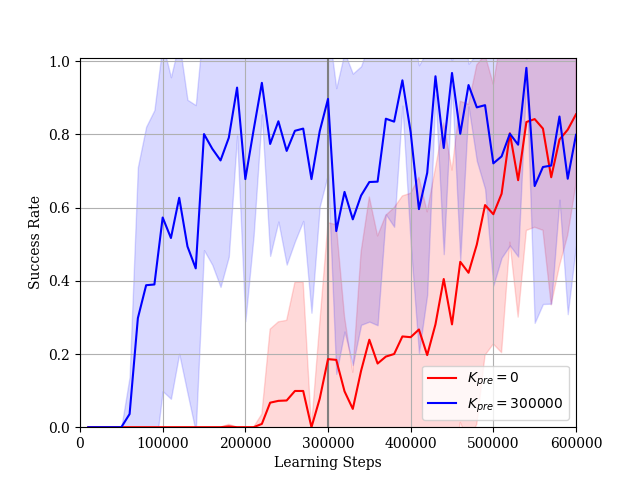}
  \caption{Success rates for the formula $\Phi_1$. The red and blue curves show the results of $K_{pre}=0$ (\textbf{Case 1}) and $K_{pre}=300000$ (\textbf{Case 2}), respectively. The solid curves and the shades represent the average results and standard deviations over $10$ trials with different random seeds, respectively. The gray line shows $300000$ steps.}
  \label{Phi_1_success_rates}
\end{center}
\end{figure}

We consider the case where the constraint is given by (\ref{specification_1}). In this simulation, we set $K=1000$ and $l_{STL}=-40$. The dimension of the extended state $z$ is $\tau=100$. The reward learning curves and the STL-rewards learning curves are shown in Figs.\ \ref{Phi_1_rewards} and \ref{Phi_1_stlrewards}, respectively. In \textbf{Case 1}, it takes a lot of steps to learn a policy such that the sum of STL-rewards is near the threshold $l_{STL}=-40$. The reward learning curve decreases gradually while the STL-reward curve increases. This is an effect of lacking in experience satisfying the STL formula $\Phi$. If the agent cannot satisfy the STL constraint during its explorations, the Lagrange multiplier $\kappa$ becomes large as shown in Fig.\ \ref{Phi_1_kappa}. Then, the STL term $-\kappa Q_{\theta_{s}}$ of the actor loss $J(\pi_{\theta})$ becomes larger than the other terms.  As a result, the agent updates the parameter vector $\theta_{\pi}$ considering only the STL rewards. On the other hand, in \textbf{Case 2}, the agent can obtain enough experiences satisfying the STL formula in $300000$ pre-training steps. The agent learns the policy such that the sum of the STL-rewards is near the threshold relatively quickly and fine-tunes the policy under the STL constraint after pre-training. According to the results in the both cases, our proposed method is useful to learn the optimal policy under the STL constraint. Additionally, as the sum of STL-rewards obtained by the learned policy is increasing, the success rate for the given STL formula is also increasing as shown in Fig.\ \ref{Phi_1_success_rates}. 

\subsubsection{Formula 2}
\begin{figure}[h]
\begin{center}
  \includegraphics[width=8.0cm]{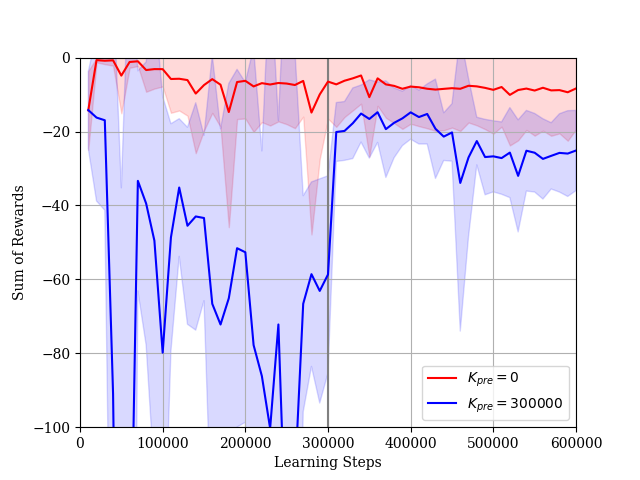}
  \caption{Reward learning curves for the formula $\Phi_2$. The red and blue curves show the results of $K_{pre}=0$ (\textbf{Case 1}) and $K_{pre}=300000$ (\textbf{Case 2}), respectively. The solid curves and the shades represent the average results and standard deviations over $10$ trials with different random seeds, respectively. The gray line shows $300000$ steps.}
  \label{Phi_2_rewards}
\end{center}
\end{figure}

\begin{figure}[h]
\begin{center}
  \includegraphics[width=8.0cm]{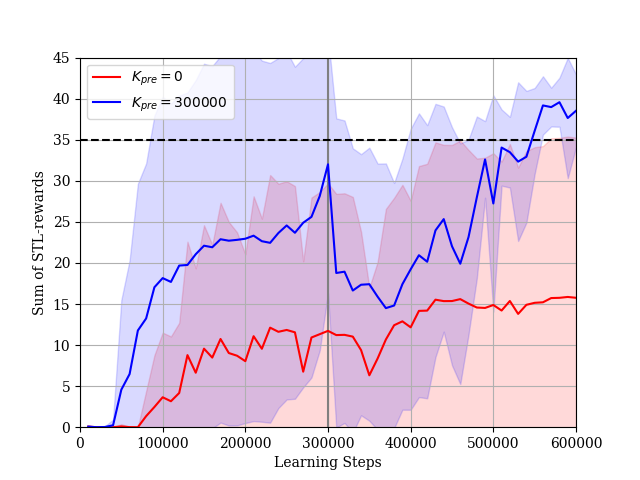}
  \caption{STL-reward learning curves for the formula $\Phi_2$. The red and blue curves show the results of $K_{pre}=0$ (\textbf{Case 1}) and $K_{pre}=300000$ (\textbf{Case 2}), respectively. The solid curves and the shades represent the average results and standard deviations over $10$ trials with different random seeds, respectively. The dashed line shows the threshold $l_{STL}=35$. The gray line shows $300000$ steps. }
  \label{Phi_2_stlrewards}
\end{center}
\end{figure}

\begin{figure}[h]
\begin{center}
  \includegraphics[width=8.0cm]{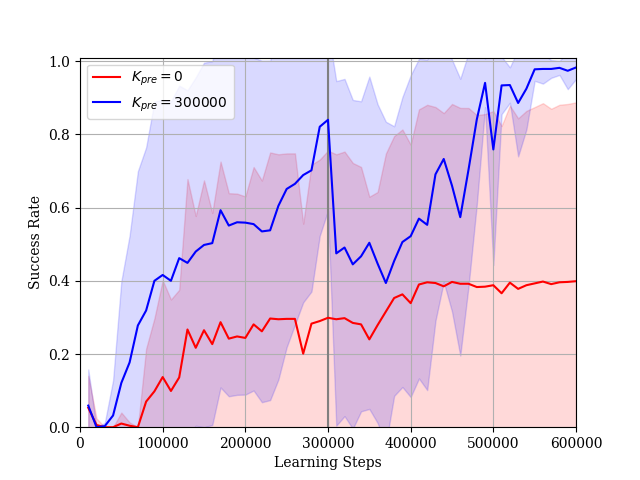}
  \caption{Success rates for the formula $\Phi_2$. The red and blue curves show the results of $K_{pre}=0$ (\textbf{Case 1}) and $K_{pre}=300000$ (\textbf{Case 2}), respectively. The solid curves and the shades represent the average results and standard deviations over $10$ trials with different random seeds, respectively. The gray line shows $300000$ steps.}
  \label{Phi_2_success_rates}
\end{center}
\end{figure}

We consider the case where the constraint is given by (\ref{specification_2}). In this simulation, we set $K=500$ and $l_{STL}=35$. The dimension of the extended state $z$ is $\tau=50$. We use the reward function $R_{STL}(z)=\exp(\beta\bm{1}(\rho(z,\phi)))/\exp(\beta)$ in stead of (\ref{tau_reward}) to prevent the sum of STL-rewards diverging to infinity. The reward learning curves and the STL-rewards learning curves are shown in Figs.\ \ref{Phi_2_rewards} and \ref{Phi_2_stlrewards}, respectively. In \textbf{Case 1}, although the reward learning curve maintains more than $-20$, the STL-reward learning curve maintains much less than the threshold $l_{STL}=35$. On the other hand, in \textbf{Case 2}, the agent learns a policy such that the sum of STL-rewards is near the threshold $l_{STL}=35$ and fine-tunes the policy under the STL constraint after pre-training. Our proposed method is useful for not only the formula $\Phi_1$ but also the formula $\Phi_2$. Additionally, as the sum of STL-rewards obtained by the learned policy is increasing, the success rate for the given STL formula is also increasing as shown in Fig.\ \ref{Phi_2_success_rates}. 

\subsection{Ablation studies for pre-processing}
\begin{figure}[h]
\begin{center}
  \includegraphics[width=8.0cm]{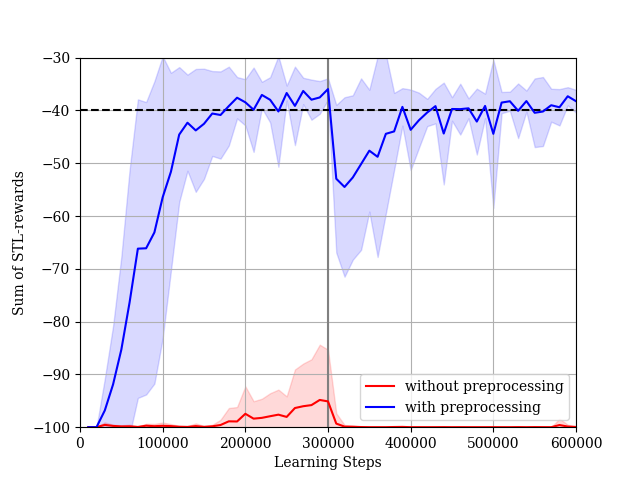}
  \caption{STL-reward learning curves for the case without pre-processing (red) and the case with pre-processing (blue). We consider the formula $\Phi_1$. The solid curves and the shades represent the average results and standard deviations over $10$ trials with different random seeds, respectively. The dashed line shows the threshold $l_{STL}=-40$. The gray line shows $300000$ steps. }
  \label{Phi_1_stlrewards_ablation}
\end{center}
\end{figure}

In this section, we show the ablation studies for pre-processing introduced in Section IV.D. We conduct the experiment for $\Phi_1$ using the SAC-Lagrangian algorithm. In the case without pre-processing, the dimensionality of the input to DNNs is $300$ and, in the case with pre-processing, the dimensionality of the input to DNNs is $5$. The STL-reward learning curves for each case are shown in Fig.\ \ref{Phi_1_stlrewards_ablation}. The agent without pre-processing cannot improve the performance of its policy for STL-rewards. The result concludes that pre-processing is useful for a problem constrained by an STL formula with a large $\tau$. 

\subsection{Comparison with another DRL algorithm}
In this section, we compare the SAC based algorithm with other algorithms: DDPG \cite{DDPG} and TD3 \cite{TD3}. TD3 is an extended DDPG algorithm with the clipped double Q-learning technique to mitigate the positive bias for the critic estimation. For the DDPG-Lagrangian algorithm and the TD3-Lagrangian algorithm, we need to set a stochastic process generating exploration noises. We use the following \textit{Ornstein-Uhlenbeck process}.
\begin{eqnarray}
\omega_{k+1}=\omega_{k}-p_1(\omega_{k}-p_2)+p_3\varepsilon,\nonumber
\end{eqnarray} 
where $\varepsilon$ is a noise generated by a standard normal distribution $\mathcal{N}(0,1)$. We set the parameters $(p_1,p_2,p_3)=(0.15,0,0.3)$. For the TD3-Lagrangian algorithm, the target policy smoothing and the delayed policy updates are same as the original paper \cite{TD3}. The target policy smoothing is implemented by adding noises sampled from the normal distribution $\mathcal{N}(0,0.2)$ to the actions chosen by the target actor DNN, clipped to $(-0.5,0.5)$, the agent updates the actor DNN and the target DNNs every $2$ learning steps. Other experimental settings such as hyper parameters, optimizers, and DNN architectures, are same as the SAC-Lagrangian algorithm.

\begin{figure}[h]
\begin{center}
  \includegraphics[width=8.0cm]{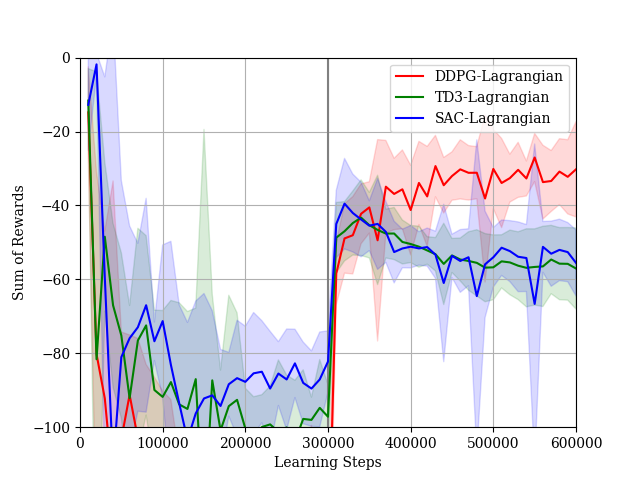}
  \caption{Reward learning curves for the formula $\Phi_1$. The red, blue, and green curves show the results of the DDPG-Lagrangian algorithm, the TD3-Lagrangian algorithm, and the SAC-Lagrangian algorithm, respectively. The solid curves and the shades represent the average results and standard deviations over $10$ trials with different random seeds, respectively. }
  \label{Phi_1_rewards_offpolicy}
\end{center}
\end{figure}

\begin{figure}[h]
\begin{center}
  \includegraphics[width=8.0cm]{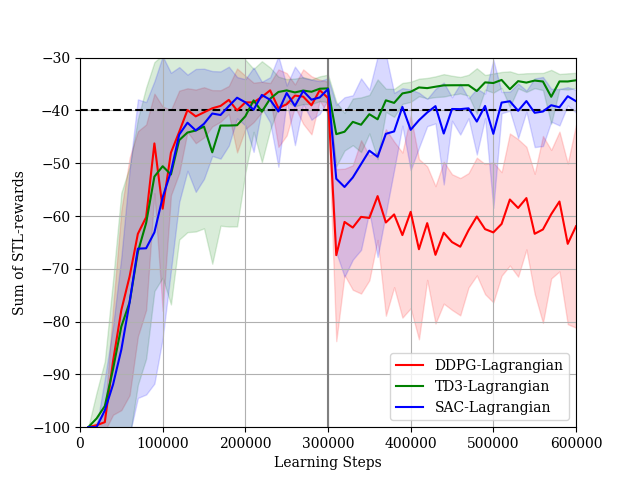}
  \caption{STL-reward learning curves for the formula $\Phi_1$. The red, blue, and green curves show the results of the DDPG-Lagrangian algorithm, the TD3-Lagrangian algorithm, and the SAC-Lagrangian algorithm, respectively. The solid curves and the shades represent the average results and standard deviations over $10$ trials with different random seeds, respectively. }
  \label{Phi_1_stlrewards_offpolicy}
\end{center}
\end{figure}

\begin{figure}[h]
\begin{center}
  \includegraphics[width=8.0cm]{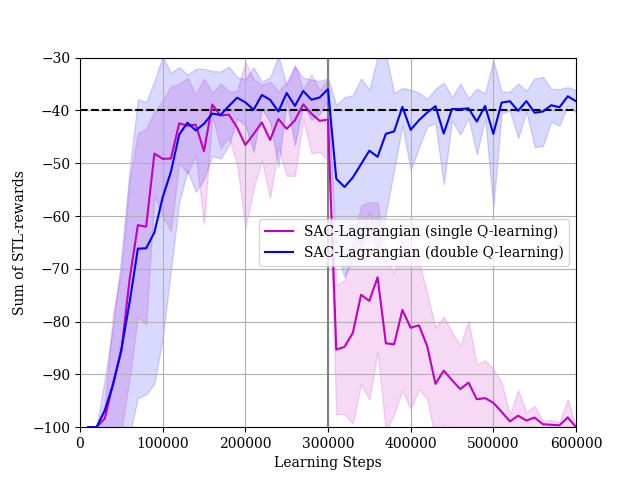}
  \caption{STL-reward learning curves for the formula $\Phi_1$. The purple and blue curves show the results of the SAC-Lagrangian algorithm without and with the double Q-learning technique, respectively.  The solid curves and the shades represent the average results and standard deviations over $10$ trials with different random seeds, respectively. }
  \label{Phi_1_stlrewards_double_Q}
\end{center}
\end{figure}

We conduct experiments for $\Phi_1$. We show the reward learning curves and the STL-reward learning curves in Figs.\ \ref{Phi_1_rewards_offpolicy} and \ref{Phi_1_stlrewards_offpolicy}, respectively. Although all algorithms can improve the policy with respect to rewards after fine-tuning, the DDPG algorithm cannot improve the policy with respect to the STL-rewards. The STL-reward curve of the DDPG-Lagrangian algorithm is much less than the threshold. On the other hand, the TD3-Lagrangian algorithm and the SAC-Lagrangian algorithm can learn the policy such that the STL-rewards are more than threshold. These results show the importance of the double Q-learning technique to mitigate positive biases for critic estimations in the fine-tuning phase. Actually, the technique is used in both the TD3-Lagrangian algorithm and the SAC-Lagrangian algorithm. Then, we show the result in the case where we do not use the double Q-learning technique in the SAC-Lagrangian in Fig.\ \ref{Phi_1_stlrewards_double_Q}. Although the agent can learn a policy such that  the STL-rewards are near the threshold in the pre-train phase, the performance of the agent's policy with respect to the STL-rewards is degraded in the fine-tune phase. 

\section{Conclusion}
We considered a model-free optimal control problem constrained by a given STL formula. We modeled the problem as a $\tau$-CMDP that is an extension of a $\tau$-MDP. To solve the $\tau$-CMDP problem with continuous state-action spaces, we proposed a CDRL algorithm with the Lagrangian relaxation. In the algorithm, we relaxed the constrained problem into an unconstrained problem to utilize a standard DRL algorithm for unconstrained problems. Additionally, we proposed a practical two-phase learning algorithm to make it easy to obtain experiences satisfying the given STL formula. Through numerical simulations, we demonstrated the performance of the proposed algorithm. First, we showed that the agent with our proposed two-phase algorithm can learn its policy for the $\tau$-CMDP problem. Next, we conducted ablation studies for pre-processing to reduce the dimensionality of the extended state and showed the usefulness. Finally, we compared three CDRL algorithms and showed the usefulness of the double Q-learning technique in the fine-tune phase.

On the other hand, the syntax in this study is restrictive compared with the general STL syntax. Relaxing the syntax restriction is a future work. Furthermore, we may not directly apply our proposed methods to high dimensional decision making problems because it is difficult to obtain experiences satisfying a given STL formula for the problems. Solving the issue is also an interesting direction for a future work.



\EOD


\begin{thebibliography}{99}

\bibitem{Sutton_RL}
 R. S. Sutton and A. G. Barto, \textit{Reinforcement Learning: An Introduction}, 2nd ed. Cambridge, MA, USA: MIT Press, 2018.

\bibitem{DRL_book}
H. Dong, Z. Ding, and S. Zhang Eds., \textit{Deep Reinforcement Learning Fundamentals, Research and Applications}, Singapore: Springer, 2020.

\bibitem{DQN}
 V. Mnih, K. Kavukcuoglu, D. Silver, A. A. Rusu, J. Veness, M. G. Bellemare, A. Graves, M. Riedmiller, A. K. Fidjeland, G. Ostrovski, S. Petersen, C. Beattie, A. Sadik, I. Antonoglou, H. King, D. Kumaran, D. Wierstra, S. Legg, and D. Hassabis, ``Human-level control through deep reinforcement learning,'' \textit{Nature}, vol. 518, pp. 529--533, Feb.\ 2015.
 
\bibitem{DRL_manipulation}
S. Gu, E. Holly, T. Lillicrap, and S. Levine, ``Deep reinforcement learning for robotic manipulation with asynchronous off-policy updates,'' in \textit{Proc.\ 2017 IEEE Int.\ Conf.\ on Robotics and Automation (ICRA)}, May 2017, pp. 3389--3396.
 
\bibitem{DRL_network}
N. C. Luong, D. T. Hoang, S. Gong, D. Niyato, P. Wang, Y.-C. Liang, and D. I. Kim, ``Applications of deep reinforcement learning in communications and networking: A survey,'' \textit{IEEE Communications Surveys \& Tutorials}, vol.\ 21, no.\ 4, pp.\ 3133--3174, May 2019.

\bibitem{DRL_multi_agent}
T. T. Nguyen, N. D. Nguyen, and S. Nahavandi, ``Deep reinforcement learning for multiagent systems: A review of challenges, solutions, and applications,'' \textit{IEEE Transactions on Cybernetics}, vol.\ 50, no.\ 9, pp.\ 3826--3839, Sept. 2020.

\bibitem{tl_control_system}
C. Belta, B. Yordanov, and E. A. Gol, \textit{Formal Methods for DiscreteTime Dynamical Systems}, Cham, Switzerland: Springer, 2017.

\bibitem{TL}
C. Baier and J.-P. Katoen, \textit{Principles of Model Checking}, Cambridge, MA, USA: MIT Press, 2008.

\bibitem{Hasanbeig}
M. Hasanbeig, A. Abate, and D. Kroening, ``Logically-Constrained Reinforcement Learning,'' 2018, \textit{arXiv:1801.08099}.

\bibitem{Yuan} 
L. Z. Yuan, M. Hasanbeig, A. Abate, and D. Kroening, ``Modular deep reinforcement learning with temporal logic specifications,'' 2019, \textit{arXiv:1909.11591}.

\bibitem{Cai}
M. Cai, M. Hasanbeig, S. Xiao, A. Abate, and Z. Kan, ``Modular deep reinforcement learning for continuous motion planning with temporal logic,'' \textit{IEEE Robotics and Automation Letters}, vol.\ 6, no.\ 4, pp.\ 7973-7980, Aug.\ 2021.

\bibitem{STL} 
O. Maler and D. Nickovic, ``Monitoring temporal properties of continuous signals,'' \textit{Formal Techniques}, \textit{Modeling and Analysis of Timed and Fault-Tolerant Systems}, pp.\ 152--166, Jan. 2004.

\bibitem{STL_Robustness}
G. E. Fainekos and G. J. Pappas, ``Robustness of temporal logic specifications for continuous-time signals,'' \textit{Theoretical Computer Science}, vol.\ 410, no.\ 42, pp.\ 4262--4291, Sept.\ 2009.

\bibitem{MPC_STL}
V. Raman, A. Donz\'{e}, M. Maasoumy, R. M. Murray, A. Sangiovanni-Vincentelli, and S. A. Seshia, ``Model predictive control with signal temporal logic specifications,'' in \textit{Proc.\ IEEE 53rd Conf.\ on Decision and Control (CDC)}, Dec.\ 2014, pp. 81--87.

\bibitem{CBF_STL}
L. Lindemann and D. V. Dimarogonas, ``Control barrier functions for signal temporal logic tasks,'' \textit{IEEE Control Systems Letters}, vol.\ 3, no.\ 1, pp.\ 96--101, Jan.\ 2019.

\bibitem{Aksaray}
D. Aksaray, A. Jones, Z. Kong, M. Schwager, and C. Belta, ``Q-learning for robust satisfaction of signal temporal logic specifications,'' in \textit{Proc.\ IEEE 55th Conf.\ on Decision and Control (CDC)}, Dec.\ 2016, pp. 6565--6570.

\bibitem{Venkataraman}
H. Venkataraman, D. Aksaray, and P. Seiler, ``Tractable reinforcement learning of signal temporal logic objectives,'' 2020, \textit{arXiv:2001.09467}.

\bibitem{Balakrishnan}
A. Balakrishnan and J. V. Deshmukh, ``Structured reward shaping using signal temporal logic specifications,'' in \textit{Proc. 2019 IEEE/RSJ Int.\ Conf.\ on Intelligent Robots and Systems (IROS)}, Nov.\ 2019, pp.\ 3481--3486.

\bibitem{Ikemoto_stl}
J. Ikemoto and T. Ushio, ``Deep reinforcement learning based networked control with network delays for signal temporal logic specifications,'' in \textit{Proc.\ IEEE 27th Int.\ Conf.\ on Emerging Technologies and Factory Automation (ETFA)}, Sept.\ 2021, doi: 10.1109/ETFA52439.2022.9921505.

\bibitem{DDPG}
T. P. Lillicrap, J. J. Hunt, A. Pritzel, N. Heess, T. Erez, Y. Tassa, D. Silver, and D. Wierstra, ``Continuous control with deep reinforcement learning,'' 2015, \textit{arXiv:1509.02971}.

\bibitem{SAC}
 T. Haarnoja, A. Zhou, K. Hartikainen, G. Tucker, S. Ha, J. Tan, V. Kumar, H. Zhu, A. Gupta, P. Abbeel, and S. Levine, ``Soft actor-critic algorithms and applications,'' 2018, \textit{arXiv:1812.05905}.

\bibitem{CMDP}
E. Altman, \textit{Constrained Markov Decision Processes}, New York, USA: Routledge, 1999.

\bibitem{Model_free_CRL}
Y. Liu, A. Halev, and X. Liu, ``Policy learning with constraints in model-free reinforcement learning: A survey,'' in \textit{Proc.\ Int.\ Joint Conf.\ on Artificial Intelligence Organization (IJCAI)}, Aug.\ 2021, pp.\ 4508--4515.

\bibitem{Kalagarla} K. C. Kalagarla, R. Jain, and P. Nuzzo, ``Model-free reinforcement learning for optimal control of Markov decision processes under signal temporal logic specifications,'' in \textit{Proc.\ IEEE 60th Conf. on Decision and Control (CDC)}, Dec.\ 2021, pp.\ 2252--2257.

\bibitem{Demo_stl_1}
A.G. Puranic, J.V. Deshmukh, and S. Nikolaidis, ``Learning from demonstrations using signal temporal logic,'' 2021, \textit{arXiv:2102.07730}.

\bibitem{Demo_stl_2}
A.G. Puranic, J.V. Deshmukh, and S. Nikolaidis, ``Learning from demonstrations using signal temporal logic in stochastic and continuous domains,'' \textit{IEEE Robotics and Automation Letters}, vol.\ 6, no.\ 4, pp. 6250-6257, Oct.\ 2021.

\bibitem{Constrained_Optimization}
D. P. Bertsekas, \textit{Constrained Optimization and Lagrange Multiplier Methods}, New York, NY, USA: Academic Press, 2014. 

\bibitem{SAC_Lagrangian}
S. Ha, P. Xu, Z. Tan, S. Levine, and J. Tan, ``Learning to walk in the real world with minimal human effort,'' 2020, \textit{arXiv:2002.08550}.

\bibitem{CSAC}
W. Wang, N. Yu, Y. Gao, and J. Shi, ``Safe off-policy deep reinforcement learning algorithm for Volt-VAR control in power distribution systems,'' \textit{IEEE Transactions on Smart Grid}, vol.\ 11, no.\ 4, pp.\ 3008--3018, Jul.\ 2020.

\bibitem{VAE}
D. P. Kingma and M. Welling, ``Auto-encoding variational Bayes,'' 2013, \textit{arXiv:1312.6114}.

\bibitem{TD3}
S. Fujimoto, H. van Hoof, and D. Meger, ``Addressing function approximation error in actor-critic methods,'' in \textit{Proc. Int.\ Conf.\ on machine learning (ICML)}, Jul.\ 2018, pp.\ 1587--1596.

\bibitem{Adam}
D. P. Kingma and J. Ba, ``Adam: A method for stochastic optimization,'' 2014, \textit{arXiv:1412.6980}.

\end{thebibliography}
\end{document}